\documentclass{article}

\PassOptionsToPackage{numbers, compress}{natbib}

\usepackage[preprint]{neurips_2021_ml4ad}

\usepackage[utf8]{inputenc} 
\usepackage[T1]{fontenc}    
\usepackage{hyperref}       
\usepackage{url}            
\usepackage{booktabs}       
\usepackage{amsfonts}       
\usepackage{nicefrac}       
\usepackage{microtype}      
\usepackage{xcolor}         
\usepackage[pdftex]{graphicx}
\usepackage[caption=false]{subfig}
\usepackage{wrapfig}
\usepackage{multirow}
\usepackage{amsmath}
\newcommand{\etal}{\textit{et al.}}

\title{Meta Guided Metric Learner for Overcoming Class Confusion in Few-Shot Road Object Detection}

\author{%
  Anay Majee \qquad \qquad
  Anbumani Subramanian \\
  Intel Corporation \\
  \texttt{firstname.lastname@intel.com} \\
  \And
  Kshitij Agrawal \\
  Bundl Technologies \\
  \texttt{kshitij.agrawal@swiggy.in} \\
}

\begin{document}

\maketitle

\begin{abstract}
    Localization and recognition of less-occurring road objects have been a challenge in autonomous driving applications due to the scarcity of data samples. Few-Shot Object Detection techniques extend the knowledge from existing base object classes to learn novel road objects given few training examples. Popular techniques in FSOD adopt either meta or metric learning techniques which are prone to class confusion and base class forgetting. In this work, we introduce a novel Meta Guided Metric Learner (MGML) to overcome class confusion in FSOD. We re-weight the features of the novel classes higher than the base classes through a novel Squeeze and Excite module and encourage the learning of truly discriminative class-specific features by applying an Orthogonality Constraint to the meta learner. Our method outperforms State-of-the-Art (SoTA) approaches in FSOD on the India Driving Dataset (IDD) by upto 11 $mAP$ points while suffering from the least class confusion of 20\% given only 10 examples of each novel road object. We further show similar improvements on the few-shot splits of PASCAL VOC dataset where we outperform SoTA approaches by upto 5.8 $mAP$ accross all splits.
\end{abstract}

\section{Introduction}
\label{intro}
Few-Shot Learning is the ability of Machine Learning models to learn novel concepts from limited training samples \cite{closerfewshot}. 
This form of learning has demonstrated its potential to alleviate the requirement of large-scale annotated datasets \cite{imagenet,coco} during model training, which is cumbersome and expensive to obtain. It also has significant importance in real-world scenarios such as autonomous navigation in unconstrained environments to detect less-occurring road objects from few-shot data samples. 
Unlike standard driving datasets \cite{kitti,city}, India Driving Dataset (IDD) \cite{idd} exhibits a real-world class-imbalanced setting and contains a set of object categories with very few annotated samples \cite{majee2021fewshot} - \emph{street carts}, \emph{water tankers}, \emph{tractors} and \emph{excavators}. Approaches tasked to learn from such real-world datasets perform poorly on the less-occurring (\emph{few-shot}) classes.

\begin{figure}[h]
      \centering
      \includegraphics[width=\textwidth]{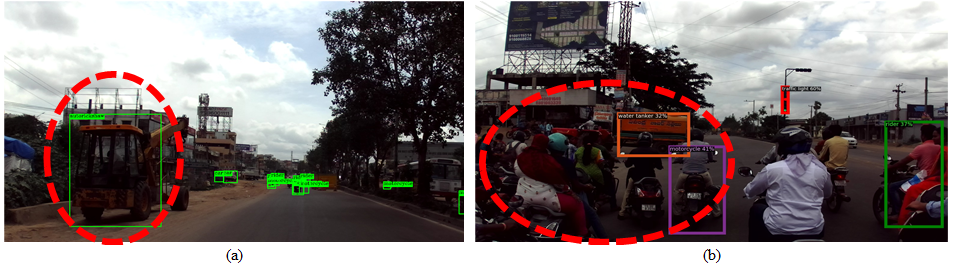}
      \caption{Example results from existing Few-Shot Object Detection technique FsDet \cite{fsdet} showing the key challenges in few-shot road object Detection. Regions marked in \textcolor{red}{red} show (a) class confusion where the \emph{Excavator} is misclassified as an \emph{Autorickshaw} and (b) catastrophic forgetting, where already learnt (base) classes are lost after few-shot adaptation.}
      \label{fig:fsod_issues}
\end{figure}
Recent developments in machine learning research have shown commendable progress in few-shot learning by extending the capability of existing models trained on large-scale datasets to adapt to sparse data. Although these models show exemplary performance for image recognition \cite{maml, nonforget, local_desc} tasks, Few-Shot Object Detection (FSOD) emerges as a relatively unexplored and complex field as it encompasses both localization and recognition tasks. 
Traditional approaches in FSOD adopt meta-learning \cite{metarcnn,addfeat} which decomposes the few-shot learning task into multiple subtasks (episodes) and aggregates their learnings through a global objective function \cite{maml,matching-net}. Recent State-of-The-Art (SoTA) approaches have adopted a simpler strategy - metric learning \cite{fsdet,fscontrastive}, which rapidly adapts to newly introduced (novel) classes by learning discriminative class boundaries between object classes.
Despite the recent successes, SoTA meta and metric learners suffer from \emph{catastrophic forgetting} and \emph{class confusion} as shown in Figure \ref{fig:fsod_issues}. This results in loss of information for the already learnt (base) classes and poor performance on the newly identified (novel) classes. 
We adopt the problem of FSOD in a real-world class-imbalanced setting to detect less-occurring road objects given a few training samples and significantly reduce the impact of class confusion on the performance of base and novel road objects. 

Unlike traditional approaches in FSOD, we propose a novel Meta Guided Metric Learning (MGML) strategy which learns class-specific feature sets through a meta learner and guides a metric learner to eliminate overlapping features between object classes. From our experiments, we notice that a large portion of low-level features is shared between the base and novel classes which when eliminated by the metric learner renders the model ineffective against intra-class variance and inter-class bias. To handle this pitfall the MGML approach introduces a Split-and-Excite module which re-weights the contribution of novel class features significantly higher than the base classes in the predictor head of the few-shot detector. We also apply an orthogonality constraint in the meta learner to encourage the learning of highly discriminative feature sets for each road object.
Unlike existing approaches in FSOD which demonstrate their performance on cannonical datasets like PASCAL VOC \cite{voc} we adopt the few-shot splits in the challenging IDD-Detection dataset \cite{majee2021fewshot} which represents a real-world, class-imbalanced setting. We show that our proposed method overcomes the large inter-class bias and intra-class variance in IDD, and suffers from the least class confusion.
The main contributions of our approach can be summarized as:
\begin{itemize}
    \item We introduce a novel Meta Guided Metric Learning approach (MGML) which combines both meta and metric learning objectives to overcome class confusion in FSOD.
    \item We apply an Orthogonality constraint (OC) and re-weight the contribution of the novel class features relatively higher than the base classes through the Split and Excite (SE) module to encourage learning of truly discriminative class-specific features during model training.
    \item We outperform SoTA FSOD approaches by upto 11 $mAP$ points while sufferering from the least class confusion of 20.12\% on the open-set of the India Driving Dataset (IDD) and demonstrate similar improvements on standard FSOD benchmarks like PASCAL VOC.
\end{itemize}

\section{Related Work}
\label{rel}
\subsection{Few-Shot Learning}
\label{rel_work:cls}
Learning algorithms in few-shot learning can be divided broadly into two categories : \emph{Metric Learning} and \emph{Meta-Learning}.
Metric learners \cite{matching-net, protonet, relation-net} learn generalizable feature representations from few-shot data, which are used to make predictions on novel tasks. A characteristic property of this class of algorithms is the use of distance / similarity metrices like cosine-similarity \cite{closerfewshot, matching-net}, euclidean distance \cite{protonet} and graph distance \cite{fsgraph} to adapt to novel classes.
Meta-Learners \cite{maml, mann} differ from metric learners by the mechanism of encoding the knowledge from few-shot data and propagating it to novel classes. Meta-learners can be further classified into memory based \cite{metalstm}, model based \cite{mann} and optimization based \cite{maml, fomaml} techniques.
Recent works \cite{mm_memory, mm_optim} adopt an architecture that combines both metric and meta learning techniques to adapt to novel classes. Our work demonstrates the effectiveness of this technique for object detection task.

\subsection{Few-Shot Object Detection}
\label{rel_work:det}

Traditional FSOD techniques \cite{lstd} adopt transfer learning to adapt to novel classes but suffer from model overfitting and catastrophic forgetting. Metric learning techniques \cite{repmet, fsdet, pnpdet} use distance metrices to rapidly adapt to novel classes. FsDet \cite{fsdet} adopts a cosine-similarity based classifier, while PNPDet \cite{pnpdet} decouples the base and novel class predictors and learns a cosine-similarity based loss function to reduce model overfitting and class confusion. Another promising direction in FSOD is the use of meta-learning techniques in conjunction with standard object detectors. Techniques like Meta-Reweight \cite{reweight}, Meta-RCNN \cite{metarcnn}, CME \cite{cme} and Add-Info \cite{addfeat} adopt this technique to learn class-specific feature sets to differentiate between base and novel class features. Fan \etal{} \cite{fsod} learns an Attention-RPN along with a relation network \cite{relation-net} to learn truly discriminative class-specific features to guide the predictor head of the object detector. Modern techniques use vision transformers \cite{metadetr} and contrastive learning \cite{fscontrastive} to improve performance on novel classes. Li \etal{} \cite{mmfsod} combines meta and metric learning techniques by adopting a pearson's distance based metric learner alongside the meta-learner.

FSOD has been recently applied in the context of autonomous driving in \cite{majee2021fewshot} to detect less-occuring road objects. The authors in \cite{majee2021fewshot} have identified \emph{class confusion} and \emph{catastrophic forgetting} as dominant roadblocks in achieving SoTA performance for road object detection. Our work adopts this problem definition and shows that a combined meta and metric learner can overcome the issue of class confusion while improving performance on novel classes.

\section{Method}
\label{meth}
\subsection{Problem Definition}
\label{meth:prob_def}
\begin{figure}[t]
      \centering
      \includegraphics[width=\textwidth]{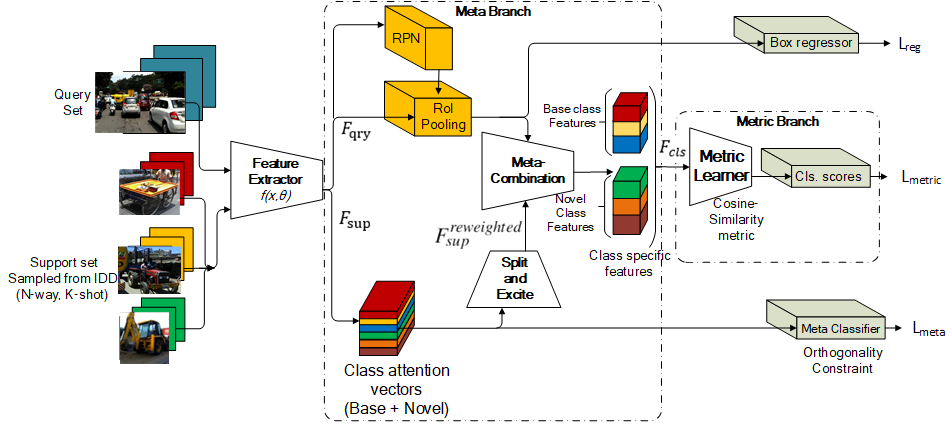}
      \caption{\textbf{Architecture of the proposed Meta Guided Metric Learner (MGML) :} Our approach employs a meta learner which guides a metric learner to learn truly discriminative features from the input dataset to adapt to novel classes.}
      \label{fig:mgml_diag}
\end{figure}
We define a proposal-based few-shot detector $h({I, \theta})$, where $I$ refers to the input data to $h(I,\theta)$ and $\theta$ represents the model parameters.
We follow the definition of a meta learner as in \cite{majee2021fewshot} and train $h(I,\theta)$ in two distinct stages - \emph{base training} and \emph{few-shot adaptation}. $h(I, \theta)$ adopts an episodic \cite{matching-net,maml} training strategy where each episode samples a subset of $N$ classes from the input dataset $D$ (\emph{$D_{base}$} during base training and \emph{$D_{base} \cup D_{novel}$} during few-shot adaptation) with $K$ examples per class, referred to as \emph{support set} and $Q$ examples ($Q$ > $K$) from $D$ containing $N$ classes, referred to as \emph{query set}. The objective of the few-shot learner $h(I, \theta)$ is to learn generalizable features from abundant training samples in $D_{base}$ during base training and rapidly adapt to novel classes in $D_{base} \cup D_{novel}$ during the few-shot adaptation stage given only $K$ examples for each class.

\subsection{Meta Guided Metric Learner}
\label{meth:mgml}
In this work, we introduce a novel Meta Guided Metric Learner (MGML) which promotes knowledge retention in base classes by learning class-specific feature representations in $D_{base} \cup D_{novel}$ and discriminates between classes by increasing the angular separation between feature clusters. 

Unlike traditional meta or metric learning architectures our proposed MGML amalgamates the benefits of both learning strategies to significantly reduce class confusion without any further impact on catastrophic forgetting. As shown in Figure \ref{fig:mgml_diag} the MGML architecture can be decomposed into two sequential branches - \emph{meta branch} and \emph{metric branch}. The meta branch learns class attentive vectors ($F_{sup}$) and class-agnostic features ($F_{qry}$) from the support and query sets in each episode. It further follows the Meta-combination module described in \cite{addfeat} to produce attentive feature sets ($F_{cls}$) for each road object. Our experiments show significant overlaps among feature sets in $F_{cls}$ which can be attributed to overlapping features in $F_{sup}$. We mitigate this problem by applying a novel Orthogonality Constraint (OC) described in section \ref{meth:oc_constraint}. 

The metric branch learns distinguishable feature representations for each class by maximizing the class boundaries through a non-linear similarity metric \cite{nonforget}. In this work we adopt a cosine similarity metric as in \cite{fsdet} to minimize the similarity between class-specific feature sets $F_{cls}$ learnt by the meta branch through a metric loss $L_{metric}$.
Despite a significant reduction in confusion, introduction of a metric learner results in a drop in novel class performance especially for classes with strong visual similarities with the base classes. We can attribute this to the elimination of distinguishable features for the confusing novel class objects by the metric learner. To mitigate this pitfall we re-weight the contribution of the novel class attentive vectors significantly higher than the base classes through the novel Split and Excite (SE) module. We describe this in detail in section \ref{meth:split_excite}.

The combined effect of the meta and metric learning objectives along with the SE and OC modules demonstrates significant reduction in class confusion while boosting the performance on novel classes in road object detection tasks \cite{majee2021fewshot}.

\subsubsection{Orthogonality Constraint}
\label{meth:oc_constraint}
For accurate classification of the class specific features in $F_{cls}$ the model $h(I, \theta)$ must learn the most discriminative feature set $F_{sup}$ which uniquely identifies each road object in the input dataset.
While standard Cross-Entropy (CE) loss function reduces the likelihood of features belonging to the same class to be closer in the feature space it does not ensure sufficient angular separation between features from different classes. This is important in few-shot road object detection due to the sparse feature sets learnt from few-shot data and large visual similarity between road objects. Ranasinghe \etal \cite{opl} imposes orthogonality in feature space for the classification task. We apply a modified orthogonality constraint more suited to few-shot detection task.

The support set in each training episode consists of K examples from N classes in the input dataset $D$. 
Each example $\{x_{i}, y_{i}\}_{i=1}^{K} \in D$ generates a class attentive vector $F_{sup_{x_{i}}} = f(x_{i}, \theta | y_{i})$ where $x_{i}$ and $y_{i}$ represents the input image and ground truth label and, $f$ is the feature extractor in the meta branch. The orthogonality constraint $L_{oc}$ maximizes the angular separation between vectors from dissimilar classes and minimizes the separation between similar ones. The computation of $L_{oc}$ is described in equation \ref{eq:loc} where the angular separation between vectors is calculated using a cosine-similarity operator. 
\begin{equation}
    L_{oc} = \sum_{\substack{i,j \in (N \times K) \\ y_{i} = y_{j}, y_{i} \neq \text{BG}}} 1 - cos(F_{sup_{x_i}}, F_{sup_{x_j}}) + \sum_{\substack{i,j \in (N \times K)\\ y_{i} \neq y_{j}, y_{i} \neq \text{BG}}} cos(F_{sup_{x_i}}, F_{sup_{x_j}})
    \label{eq:loc}
\end{equation}

$L_{oc}$ is applied as an additional loss term to the CE loss $L_{ce}$ in the objective function of the meta branch $L_{meta}$ as shown in equation \ref{eq:lmeta}. $L_{oc}$ is applied only to the foreground classes as the background (BG) class can potentially contain information from multiple object classes. The hyperparameter $\alpha$ controls the contribution of the orthogonality constraint in $L_{meta}$ and is described in section \ref{abl:params}.
\begin{equation}
    L_{meta} = L_{ce} + \alpha L_{oc}
    \label{eq:lmeta}
\end{equation}


\begin{figure}
        \centering
        \small
        \begin{tabular}{lccc}
            \rotatebox{90}{FsDet \cite{fsdet}} &
                \includegraphics[width=0.3\textwidth]{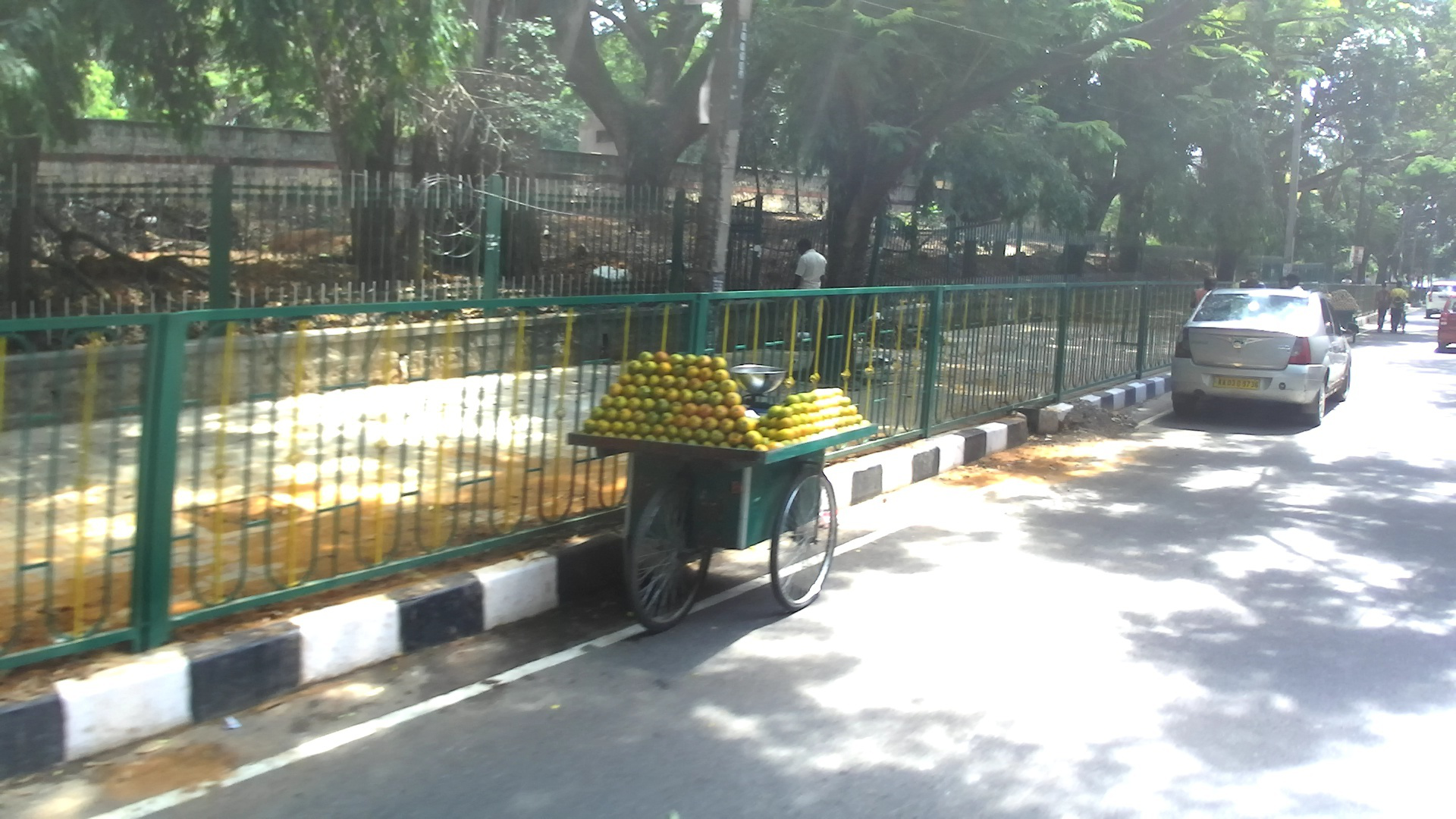} & \includegraphics[width=0.3\textwidth]{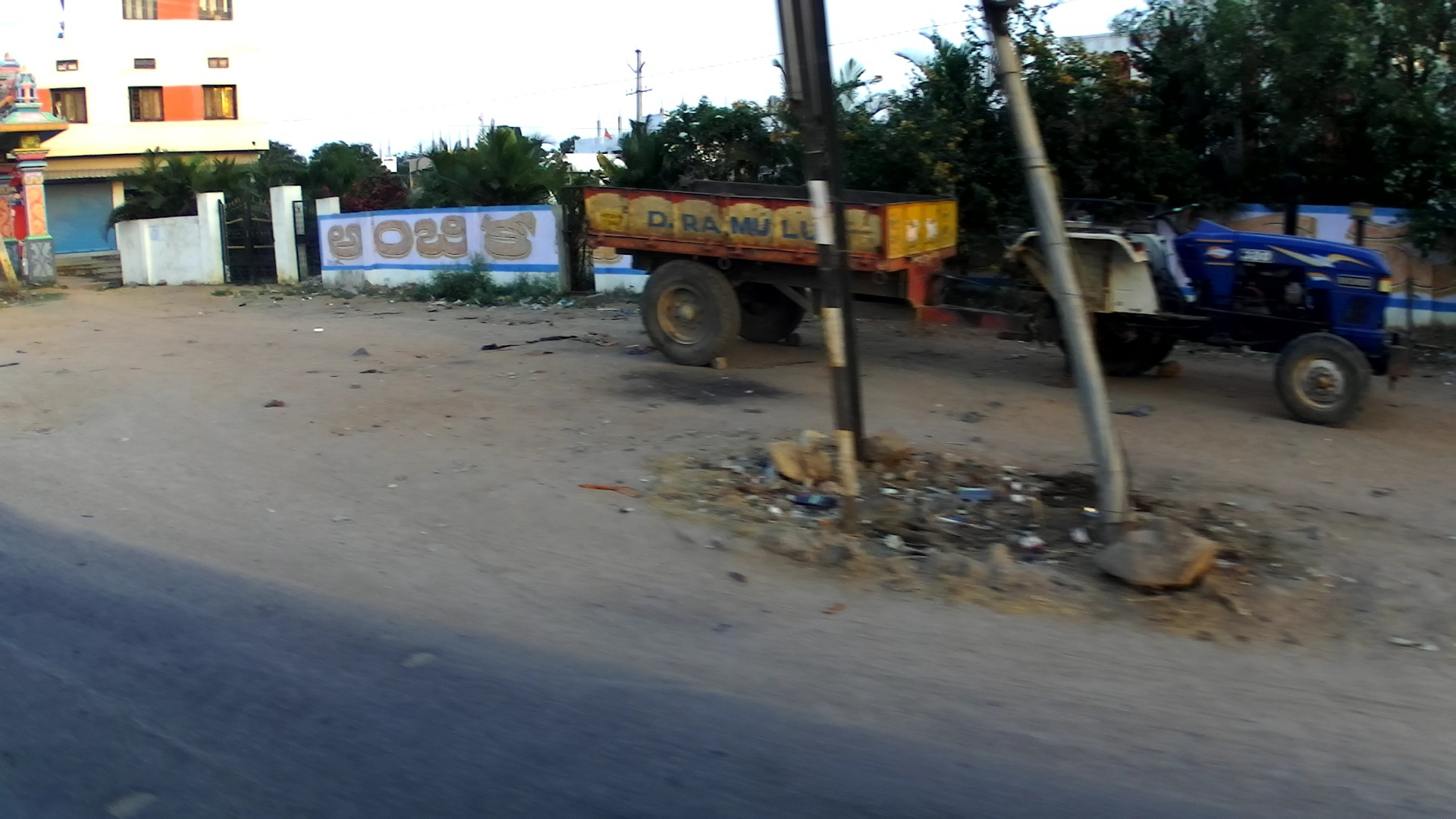} &
                \includegraphics[width=0.3\textwidth]{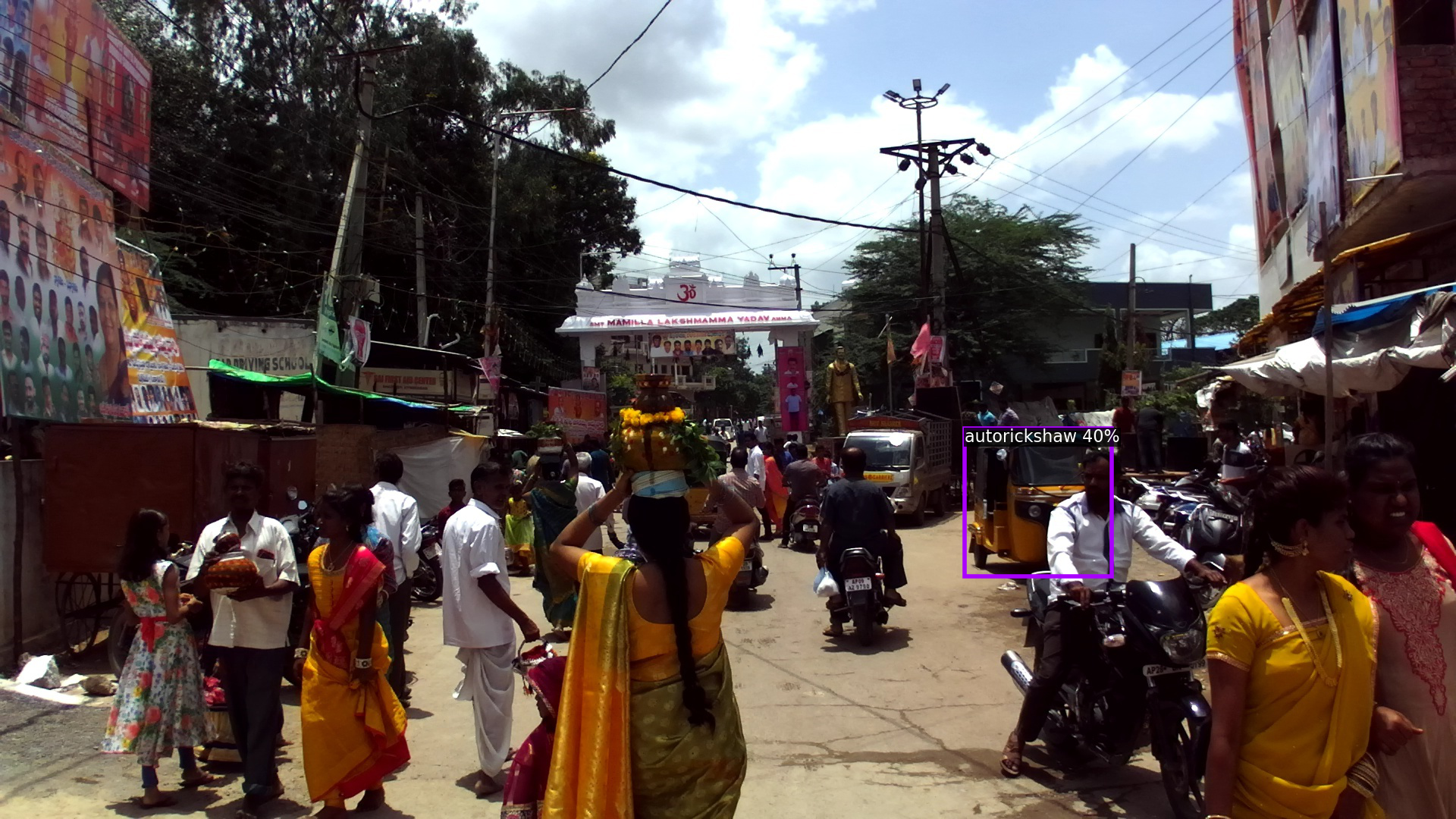} \\ 
            \rotatebox{90}{MGML (ours)} & 
                \includegraphics[width=0.3\textwidth]{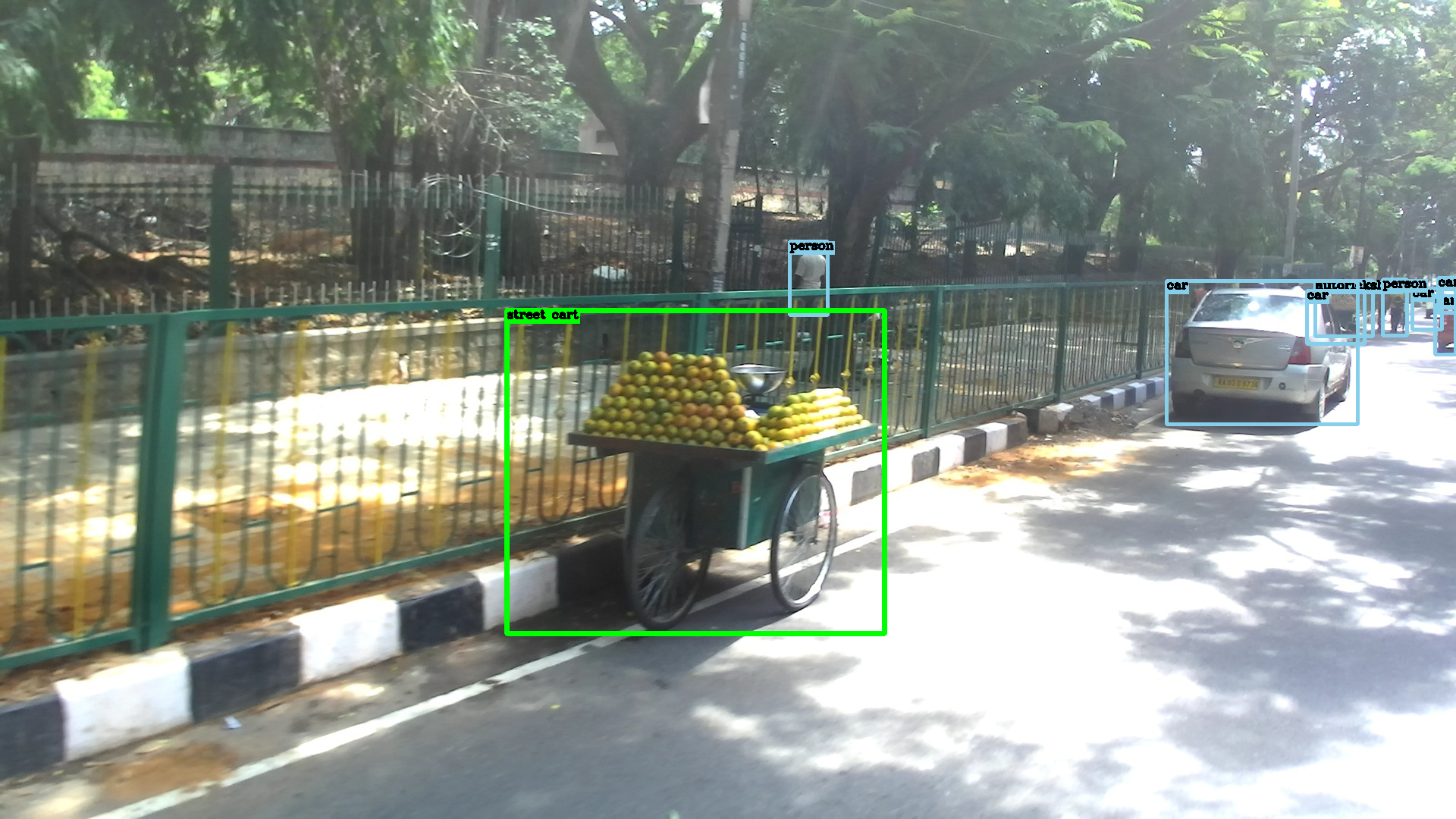} & \includegraphics[width=0.3\textwidth]{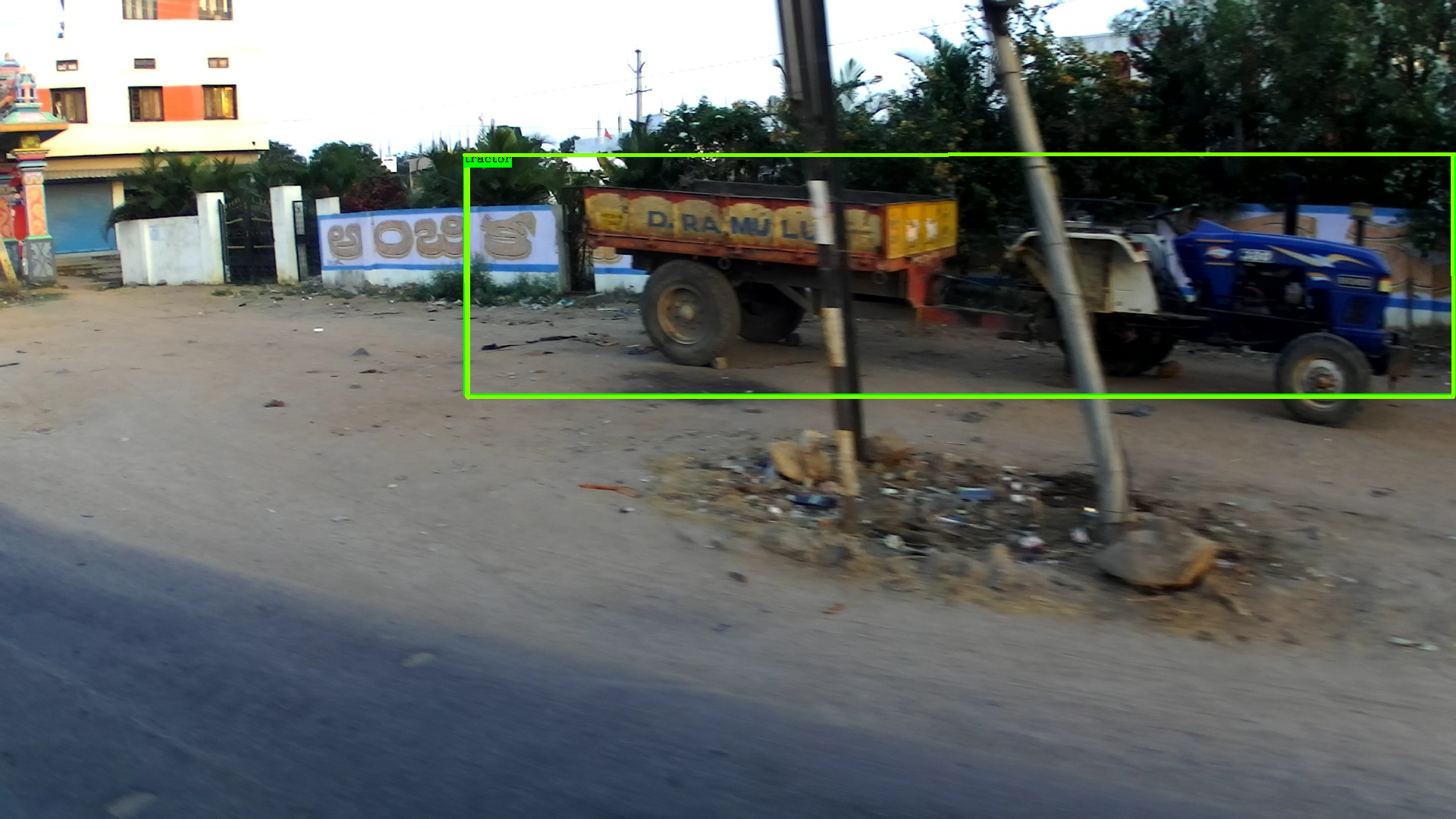} &
                \includegraphics[width=0.3\textwidth]{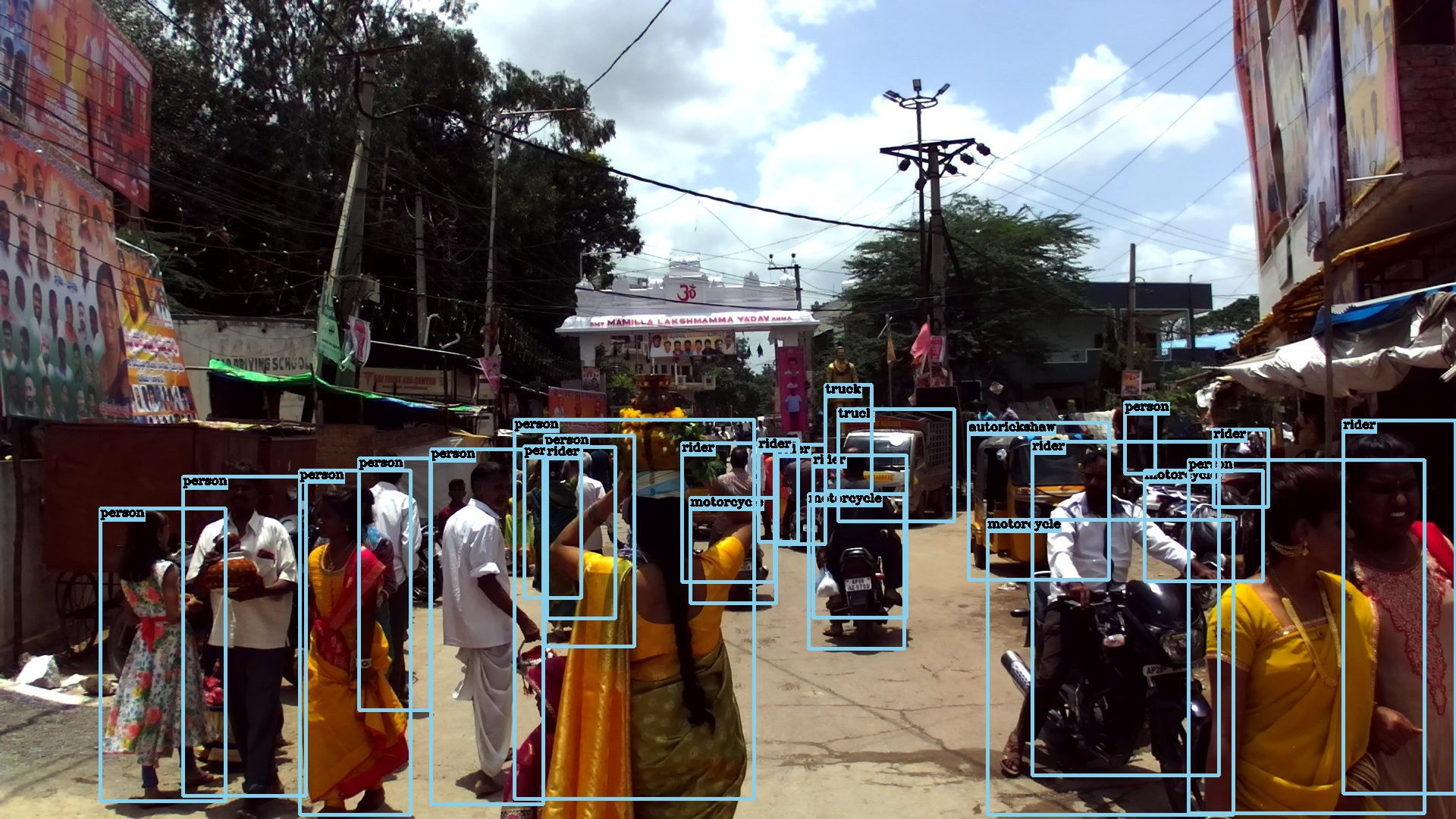} \\ 
            & (a) & (b) & (c) \\ 
        \end{tabular}
        \caption{\textbf{Qualitative results from the few-shot India Driving Dataset: } We contrast the performance of MGML against FsDet, for novel classes in the IDD-OS split for the 10-shot setting. FsDet suffers from extreme catastrophic forgetting and is unable to adapt to large intra-class and inter-class variations in IDD which are overcome by MGML.}
        \label{fig:qual}
\end{figure}

\subsubsection{Split and Excite Module}
\label{meth:split_excite}
The proposed Split and Excite (SE) module in MGML re-weights the class-specific vectors in $F_{sup}$ pertaining to the novel classes higher with respect to the base classes in the few-shot adaptation stage. This module highlights the sparse features from the novel objects and reduces the chance of feature elimination due to the addition of the metric learning objective. 
This module can be formulated as three distinct phases. At first, $F_{sup}$ is \emph{split} into base ($F^{base}_{sup}$) and novel ($F^{novel}_{sup}$) class vectors.
Secondly, the vectors from $F^{novel}_{sup}$ are re-weighted by channel-wise multiplication of a learnable hyper-parameter $\lambda$ (\emph{excite} phase). 
Finally, the base and novel feature vectors are aggregated with the class-agnostic query set features to form $F^{reweighted}_{cls_i}$ as shown in equation \ref{eq:freweighted}. We follow the aggregation process described in \cite{addfeat} to produce the class-specific feature set $F_{cls_i}$ for each class $i$ in the input dataset as shown in equation \ref{eq:fcls}.
\begin{equation}
    F^{reweighted}_{cls_i} = [F_{qry} \otimes F^{base}_{sup}, F_{qry} \otimes ( \lambda \ast F^{novel}_{sup})],
    \label{eq:freweighted}
\end{equation}
\begin{equation}
    F_{cls_i} = [F^{reweighted}_{cls_i}, F_{qry} - F_{sup},   F_{qry}]
    \label{eq:fcls}
\end{equation}
where $N$ represents the total number of classes during the few-shot adaptation stage. 
Through our experiments, we empirically show that such a formulation helps to further boost the performance on novel classes without any degradation in the base classes.

\subsubsection{Training Methodology}
As described in section \ref{meth:prob_def}, $h(I, \theta)$ is trained in two stages. During the base training stage $h(I, \theta)$ is trained on abundant samples in $D_{base}$ till convergence by adopting the meta training strategy in \cite{metarcnn} and applying the orthogonality constraint ($L_{meta}$) to the meta branch. We use the loss functions in \cite{faster-rcnn} comprising of a binary Cross-Entropy (CE) loss at the Region Proposal Network (RPN) to separate foreground and background proposals $L_{rpn}$, a cross-entropy loss for bounding box classifier $L_{cls}$ and a smoothed L1 loss to localize the bounding box deltas $L_{reg}$. During the few-shot adaptation stage, we introduce the metric branch into $h(I, \theta)$ and apply the SE and OC modules in the meta branch to adapt to K-shot data in $D_{base} \cup D_{novel}$. The box classification loss $L_{cls}$ is replaced with a combined meta loss $L_{meta}$ and a cosine similarity penalty $L_{metric}$ as described in equation \ref{eq:loss_total}.
\begin{equation} \label{eq:loss_total}
L = L_{meta} + L_{metric} +  L_{reg}
\end{equation}

\section{Experiments}
\label{exp}
\begin{table*}[t]
      \caption{\textbf{Results on few-shot splits of the India Driving Dataset (IDD): } Few-Shot object detection performance on novel classes in IDD-10 and IDD-OS splits from IDD for 5 and 10-shot settings. }
      \centering
      \scalebox{0.8}{
      \begin{tabular}{l|c|cc|cc|cc}
            \hline
            \multirow{2}{*}{Method}          & \multirow{2}{*}{Meta/Metic Learner} &
            \multicolumn{2}{c|}{\textbf{IDD-OS (Open-Set)}} &
            \multicolumn{2}{c|}{\textbf{IDD-10 (Split-1)}}     &   \multicolumn{2}{c}{\textbf{IDD-10 (Split-2)}}  \\ 
                &  &  \emph{K=5}      &   10  &  5     &   10 & 5   &  10\\ 
            \hline
            Meta-RCNN  \cite{metarcnn}   & Meta & 4.3  & 6.4  & 5.7  & 7.8 & 7.4  & 6.7      \\
            Add-Info \cite{addfeat} & Meta & 18.2  & 28.8  & 5.2 & 10.0  & 7.7  & 9.5    \\
            FsDet w/ cos \cite{fsdet}  & Metric & 23.6  & 39.8   & 13.1  & \textbf{22.1}  & 14.8  & \textbf{22.8}   \\
            \hline
            \textbf{Ours (MGML + SE + OC)} & Meta + Metric &  \textbf{28.0} &	\textbf{48.0} &	\textbf{15.1} &	17.2 &	\textbf{15.2}	 & 18.6 \\
            \hline
      \end{tabular}}
      \label{tab:idd_os}
\end{table*}
In this section, we describe our experimental setup and benchmark the performance of our proposed MGML technique on two few-shot object detection benchmarks - India Driving Dataset \cite{majee2021fewshot} and PASCAL-VOC dataset \cite{reweight}. For all our experiments we report the Mean Average Precision ($mAP_{50}$) at 50\% Intersection over Union (IoU) \cite{voc}, which is a standardized metric for evaluating object detection performance.

\subsection{Datasets}
\label{exp:datasplits}

\textbf{India Driving Dataset (IDD)} \cite{idd} consists of 15 object classes (in the detection dataset), representing traffic scenes on Indian roads. For the few-shot tasks we adopt the benchmark splits in \cite{majee2021fewshot}- IDD-10 and IDD-OS, which represents a real-world class imbalanced setting.

\begin{table}
    \centering
    \scalebox{0.8}{
    \begin{tabular}{l|cccc|cc}
            \hline
            \multirow{2}{*}{Method} & Meta\footnotemark & Metric  & SE Module         & Orthogonality &\multirow{2}{*}{$mAP_{base}$} & \multirow{2}{*}{$mAP_{novel}$} \\ 
                                    & Learner           & Learner & ($\lambda$=2.0)   & Constraint    &  &\\
            \hline
            Add-Info                & \checkmark    &            &              &     & 37.1  & 28.8 \\
            FsDet w/ cos            &               & \checkmark &              &     & \textbf{47.4}  & 37.0     \\
            \hline
            \textbf{\multirow{4}{*}{MGML (ours)}}    & \checkmark    & \checkmark &              &   & 38.0  & 40.0     \\
                                                     &  \checkmark   & \checkmark & \checkmark   &   & 38.0  & 45.4  \\
                                                     &  \checkmark   & \checkmark &    & \checkmark  & 41.0  & 46.1  \\
                                                     &  \checkmark   & \checkmark & \checkmark  & \checkmark  & 41.5  & \textbf{48.0}  \\
            \hline
      \end{tabular}
      }
    \caption{\textbf{Components of the proposed MGML architecture:} Highlights the contribution of each architectural block towards the base and novel class performance of the IDD-OS split in the 10-shot setting.}
    \label{tab:mgml_comp}
\end{table}
\footnotetext{Adapted from Xiao \etal{} \cite{addfeat}.}

\textbf{IDD-10} consists of 10 representative classes from IDD which are divided into 7 base classes and 3 novel classes. Based on the choice of novel classes we consider two data splits, referred to as split-1 (\emph{bicycle}, \emph{bus} and \emph{truck} as novel classes) and split-2 (\emph{autorickshaw}, \emph{motorcycle} and \emph{truck} as novel classes). 

\textbf{IDD-OS} consists of 14 classes, with 10 base classes and 4 novel classes. The 4 novel classes (\emph{Street cart}, \emph{Tractor}, \emph{Water tanker} and \emph{Excavator}) were generated by expanding on the \emph{vehicle fallback} category in IDD and represents the open world deployment setting.

We use the complete \emph{train} set of IDD for base training and sample N-way K-shot (\emph{K=5, 10}) episodes during few-shot adaptation. We use the \emph{val} set of IDD for evaluation.

\textbf{PASCAL-VOC} \cite{voc} consists of 20 object classes, from which 15 are considered as base and 5 are considered as novel classes. The 5 novel classes are randomly chosen to form 3 representative category splits. We follow the data splits from Meta-Reweight \cite{reweight} and evaluate our methods on novel split-1 (\emph{bird}, \emph{bus}, \emph{cow}, \emph{motorbike} and \emph{sofa}), novel split-2 (\emph{aeroplane}, \emph{bottle}, \emph{cow}, \emph{horse} and \emph{sofa}) and novel split-3 (\emph{boat}, \emph{cat}, \emph{motorbike}, \emph{sheep} and \emph{sofa}) for \emph{K=1, 5 and 10} shot settings. For training we use the complete \emph{trainval} split from PASCAL-VOC07+12 datasets and \emph{test} split of PASCAL-VOC 2007 for evaluation.

\begin{table*}[t]
      \caption{Few-shot object detection performance ($mAP_{novel}$) on novel class splits of PASCAL-VOC dataset. We tabulate results for K={1,5,10} shots from various SoTA techniques in FSOD. Results are averaged over 10 runs. The symbol $-$ indicates that results were not published by the authors.}
      \centering
      \small
      \scalebox{0.8}{
      \begin{tabular}{l|c|c|ccc|ccc|ccc}
            \hline
            \textbf{Method}          & \multicolumn{1}{p{2cm}|}{\centering Meta/ Metric \\ Learner} & Backbone &  
                                        \multicolumn{3}{c|}{\textbf{Novel Split 1}} &   
                                        \multicolumn{3}{c|}{\textbf{Novel Split 2}} &
                                        \multicolumn{3}{c}{\textbf{Novel Split 3}} \\ 
                                     &&&  K=1 & 5 & 10 &
                                          1   & 5 & 10 &
                                        1   & 5 & 10 \\ 
            \hline
            Meta-RCNN \cite{metarcnn}       & Meta      & FRCN-R101 & 19.9  & 45.7  & 51.5  & 10.4  & 34.8  & 45.4  & 14.3  & 41.2  & 48.1  \\
            Meta-Reweight \cite{reweight}   & Meta      & YOLO V2   & 14.8  & 33.9  & 47.2  & 15.7  & 30.1  & 40.5  & 21.3  & 42.8  & 45.9  \\
            MetaDet \cite{metadet}          & Meta      & FRCN-R101 & 18.9  & 36.8  & 49.6  & 21.8  & 31.7  & 43.0  & 20.6  & 43.9  & 44.1  \\
            Add-Info \cite{addfeat}         & Meta      & FRCN-R101 & 24.2  & 49.1  & 57.4  & 21.6  & 37.0  & 45.7  & 21.2  & 43.8  & 49.6  \\
            CME \cite{cme}            & Meta & YOLO V2 & 17.8   & 44.8  &  47.5 &  12.7 & 33.7   & 40.0   & 15.7  &  44.9 & 48.8  \\
            \hline
            PNPDet \cite{pnpdet}            & Metric    & DLA-34    & 18.2  & - & 41.0  & 16.6  & -  & 36.4  & 18.9  & -  & 36.2  \\
            FsDet w/ FC \cite{fsdet}        & Metric    & FRCN-R101 & 36.8  & 55.7  & 57.0  & 18.2  & 35.5  & 39.0  & 27.7  & 48.7  & 50.2  \\
            FsDet w/ cos \cite{fsdet}       & Metric    & FRCN-R101 & 39.8  & 55.7  & 56.0  & 23.5  & 35.1  & 39.1  & 30.8  & 49.5  & 49.8  \\
            FSCE \cite{fscontrastive}            & Metric & FRCN-R101 & \textbf{28.2} &  46.2 & 54.1 & 16.5	& 35.9 & 45.3 & 22.2	& 45.4	& 49.4 \\
            \textbf{ours (MGML + SE + OC)}         & Meta + Metric & FRCN-R101 & 27.3  & \textbf{51.4}  & \textbf{58.0}  & \textbf{22.3}  & \textbf{37.9}  & \textbf{45.6} & \textbf{22.7}  & \textbf{47.1} & \textbf{51.2}  \\
            \hline 
      \end{tabular}
      \label{tab:voc_main}}
\end{table*}

\subsection{Implementation Details}
The MGML architecture is based on the proposal-based Faster-RCNN \cite{faster-rcnn} object detector with a ResNet-101 \cite{resnet} backbone. Input to the network consists of a batch of 4 query images resized to 600 x 800 pixels and \emph{K}-shot support images for each query image, resized to 224 x 224 pixels. We consider images with object sizes greater than 100 x 100 pixels for the support set to ensure the quality of extracted features. We follow the base training procedure in \cite{addfeat} to train our model for till convergence (20 epochs). During the few-shot adaptation stage, we train our model for 12 epochs with a constant learning rate of 1 x $10^{-3}$. We use an Adam \cite{adam} optimizer and momentum value of 0.9. The hyperparameter $\lambda$ is introduced in the few-shot adaptation stage and $\alpha$ is applied in both stages of model training. The values of $\alpha$ and $\lambda$ are chosen through ablation experiments in section \ref{abl:params}. All benchmark experiments are conducted on a single GPU with 12GB memory.

\subsection{Results on India Driving Dataset}
\label{exp:idd}
We benchmark our MGML approach against the FSOD benchmark on IDD as in \cite{majee2021fewshot}.
Table \ref{tab:idd_os} records the performance of our MGML approach on IDD-10 and IDD-OS splits for 5 and 10 shot settings.
We benchmark our approach against both meta and metric learning approaches in \cite{majee2021fewshot}. Results show that our approach (MGML + SE + OC) outperforms SoTA approaches on the IDD-OS split by large margins, upto 30\% (11 $mAP$ points) across 5 and 10-shot settings. Such results establish the superiority and robustness of our approach in detecting less-occuring road objects in class imbalanced driving environments.
On the IDD-10 split, our MGML approach outperforms SoTA meta learning approaches (Add-Info) by upto 9.9 $mAP$ points but under-perform against metric learning approaches. We observe that the MGML does not achieve high performance gains for higher values of $K$ (10-shot) in IDD-10. This can be attributed to the large intra-class variance among road objects in IDD.

Figure \ref{fig:qual} demonstrates the qualitative results from our approach against the SoTA metric learner (FsDet) on the IDD-OS split for the 10-shot setting. 
We demonstrate the robustness of our MGML approach against pitfalls in SoTA FSOD techniques such as large intra-class bias (Figure \ref{fig:qual}(a),\ref{fig:qual}(b)), catastrophic forgetting (Figure \ref{fig:qual}(c)) and ineffectiveness against occlusions (Figure \ref{fig:qual}(b)).

\subsection{Results on PASCAL VOC dataset}
\label{exp:voc}
We benchmark the MGML approach against SoTA approaches on the few-shot datasplits of the PASCAL VOC dataset as in \cite{fscontrastive}. 
Following the datasplits in section \ref{exp:datasplits} we conduct our experiments on K=1,5, 10 settings and summarize the results in Table \ref{tab:voc_main}.
We show that the MGML approach outperforms SoTA approaches on almost all three splits. The maximum improvement was observed in split-1 for the 10-shot setting where MGML approach the SoTA approach (FSCE) by 7.2\% (3.9 $mAP$ points).

\section{Ablations}
\label{abl}
\subsection{Components of Our Proposed MGML Architecture}
\label{abl:comp}
\begin{wraptable}{R}{7.5cm}
\caption{Ablation for the effect of key hyper-parameters ($\lambda$ and $\alpha$) on novel class performance in IDD-OS (10-shot setting). The chosen values are \underline{underlined} and associated scores are indicated in \textbf{bold}.}
\label{tab:hyper}
\centering
\small
\begin{tabular}{c|c|c|c}
\hline
Parameter & Value & $mAP_{base}$ & $mAP_{novel}$ \\ \hline
\multirow{5}{2cm}{\centering $\lambda$ \\($\alpha$ = 0.0)}
                              & 1.0                       & 37.9                         & 40.0                          \\
                              & 1.5                       & 38.0                         & 42.8                          \\
                              & \underline{2.0}                       & \textbf{38.1}                & \textbf{45.0}                          \\
                              & 2.5                       & 37.0                         & 40.9                          \\
                              \hline
\multirow{3}{2cm}{\centering $\alpha$ \\($\lambda = 2.0$)}     
                              & 0.05                 & 40.5                         & 46.8                          \\
                              & 0.1                   & 40.8                         & 47.4                          \\
                              & \underline{0.5}                    & \textbf{41.0}                & \textbf{47.9}                          \\
                              & 1.0                   & 40.9                         & 47.0                          \\
                              & 2.0                   & 41.0                         & 45.7                          \\
                              \hline
\end{tabular}
\end{wraptable}
The MGML architecture can be decomposed into three major components. Table \ref{tab:mgml_comp} demonstrates the contribution of each component on the base and novel class performance. At first, we combine both meta and metric learners into a single unified architecture where the class specific features learnt by the meta learner guides the metric learner. 
Secondly, we introduce the SE module which re-weights the novel class attention vectors higher than base classes to reduce chances of feature elimination during few-shot adaptation.
Finally, we apply a novel Orthogonality Constrant on the meta branch to learn the most distinguishable feature vector for each class. 
The combined effect of all the three components demonstrates the best novel class performance on the IDD-OS split in the 10-shot setting but continues to suffer from base class forgetting due to large inter-class bias among road objects. 

\begin{figure}
      \centering
      \includegraphics[width=\textwidth]{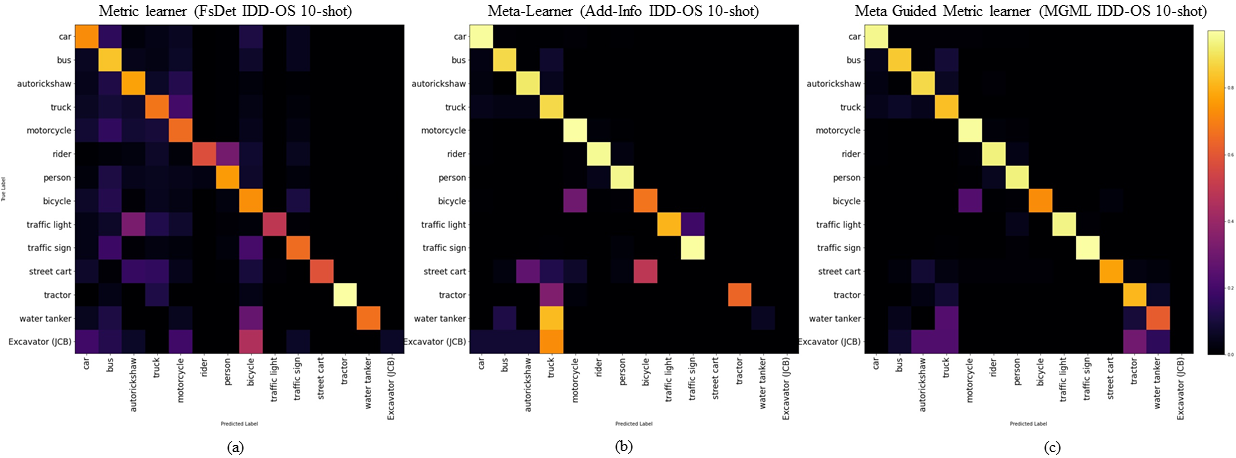}
      \caption{Confusion matrix plotted for three categories of few-shot object detection networks (a) Metric Learner (FsDet) (b) Meta-Learner (Add-Info) (c) Meta Guided Metric Learner (MGML) trained using 10-shot samples in IDD-OS split. MGML shows least class confusion of 20.1 \%.}
      \label{cls_conf:idd}
\end{figure}

\subsection{Hyper-parameters of The MGML Technique}
\label{abl:params}
The formulation of MGML introduces two hyper-parameters $\lambda$ and $\alpha$. Initially we choose the value of $\alpha = 0$ and vary the value of $\lambda$ in the range of 1.0 to 2.5. We observe an increase in novel class performance with increase in $\lambda$ between 1.0 and 2.0 and a steep reduction thereafter. Thus we chose $\lambda= 2.0$ for all our experiments.
We then fixate the value of $\lambda$ at 2.0 and vary the value of $\alpha$ in the range of [0.05, 2.0]. We observe a small improvement in novel class performance in the range of 0.05 to 0.5. We also observe a significant boost in base class performance with the introduction of the OC constraint. Based on the results in table \ref{tab:hyper} we choose the value of $\alpha$ as 0.5 for all our benchmark experiments. 


\subsection{Class Confusion Between Base and Novel Classes}
\label{abl:conf}
Class confusion stands out as a prominent issue in few-shot object detection in the context of autonomous driving as road objects share a large number of visual features \cite{majee2021fewshot} among themselves. Figure \ref{cls_conf:idd} compares class confusion of our approach (MGML) with popular FSOD architectures (FsDet and Add-Info) through confusion matrices. FsDet (metric learner) shows the highest confusion (43.9 \%) between both base and novel classes. On the other hand, Add-Info (Meta-Learner) shows reduced confusion among base classes but elevated confusion among novel classes (average confusion of 31.05 \%). Finally, MGML shows least confusion between both base and novel classes (20.12 \%), thus emerging as a technique of choice for few-shot learning in autonomous driving.

\section{Conclusion}
In this work, we introduced a novel FSOD technique - Meta Guided Metric Learner (MGML) to overcome the impeding effect of class confusion in detecting less-occurring road objects in driving environments.
Unlike existing approaches in FSOD we employ both meta and metric learning objectives in an unified proposal-based architecture.
The introduced Orthogonality Constraint and Split and Excite module ensures the learning of discriminative feature sets to overcome inter-class variance and intra-class bias among road objects.
Our approach achieves SoTA performance on the open-set split of the challenging India Driving Dataset by demonstrating upto 30\% improvement in novel class performance over existing methods in a real-world class-imbalanced setting.
Alongside improvements in absolute performance, the MGML approach suffers the least class confusion of 20.12\% among SoTA FSOD approaches.
Catastrophic forgetting of base classes continues to be an impeding issue in few-shot road object detection and will be addressed in future research.


\section{Acknowledgements}
The authors would like to thank Pravin Chandran (Intel Corporation) and Ashutosh Agarwal (IIT Delhi) for their valuable comments at various stages of this project. We would also like to thank Raghavendra Bhat (Intel Corporation) for his feedback on this paper.
\small
\bibliography{references}{}

\begin{thebibliography}{10}

\bibitem{lstd}
Hao Chen, Yali Wang, Guoyou Wang, and Yu~Qiao.
\newblock {LSTD:} {A} low-shot transfer detector for object detection.
\newblock In {\em Proc. of the Thirty-Second {AAAI} Conf. on Artificial
  Intelligence}, pages 2836--2843, 2018.

\bibitem{closerfewshot}
Wei-Yu Chen, Yen-Cheng Liu, Zsolt Kira, Yu-Chiang Wang, and Jia-Bin Huang.
\newblock A closer look at few-shot classification.
\newblock In {\em Intl. Conf. on Learning Representations}, 2019.

\bibitem{mm_memory}
Yu~Cheng, Mo~Yu, Xiaoxiao Guo, and Bowen Zhou.
\newblock Few-shot learning with meta metric learners.
\newblock {\em ArXiv}, abs/1901.09890, 2019.

\bibitem{city}
Marius Cordts, Mohamed Omran, Sebastian Ramos, Timo Rehfeld, Markus Enzweiler,
  Rodrigo Benenson, Uwe Franke, Stefan Roth, and Bernt Schiele.
\newblock The {Cityscapes} {Dataset} for semantic urban scene understanding.
\newblock In {\em IEEE Conf. on Computer Vision and Pattern Recognition
  (CVPR)}, 2016.

\bibitem{imagenet}
J.~{Deng}, W.~{Dong}, R.~{Socher}, L.~{Li}, {Kai Li}, and {Li Fei-Fei}.
\newblock {ImageNet:} {A} large-scale hierarchical image database.
\newblock In {\em IEEE Conf. on Computer Vision and Pattern Recognition
  (CVPR)}, pages 248--255, 2009.

\bibitem{voc}
M.~Everingham, L.~Van~Gool, C.~K.~I. Williams, J.~Winn, and A.~Zisserman.
\newblock The {Pascal} {Visual} {Object} classes ({VOC}) {Challenge}.
\newblock {\em Intl. Journal of Computer Vision}, pages 303--338, 2010.

\bibitem{fsod}
Qi~Fan, Wei Zhuo, Chi-Keung Tang, and Yu-Wing Tai.
\newblock Few-shot object detection with {Attention-RPN} and {Multi-Relation}
  detector.
\newblock In {\em IEEE Conf. on Computer Vision and Pattern Recognition
  (CVPR)}, 2020.

\bibitem{maml}
Chelsea Finn, Pieter Abbeel, and Sergey Levine.
\newblock {Model-Agnostic} {Meta-Learning} for fast adaptation of deep
  networks.
\newblock In {\em Proc. of the 34th Intl. Conf. on Machine Learning}, 2017.

\bibitem{kitti}
Andreas Geiger, Philip Lenz, and Raquel Urtasun.
\newblock Are we ready for {Autonomous Driving}? the {KITTI Vision Benchmark
  Suite}.
\newblock In {\em Conf. on Computer Vision and Pattern Recognition (CVPR)},
  2012.

\bibitem{nonforget}
Spyros Gidaris and Nikos Komodakis.
\newblock Dynamic few-shot visual learning without forgetting.
\newblock In {\em IEEE Conf. on Computer Vision and Pattern Recognition
  (CVPR)}, 2018.

\bibitem{resnet}
Kaiming He, Xiangyu Zhang, Shaoqing Ren, and Jian Sun.
\newblock Deep residual learning for {Image} {Recognition}.
\newblock In {\em IEEE Conf. on Computer Vision and Pattern Recognition
  (CVPR)}, 2016.

\bibitem{reweight}
Bingyi Kang, Zhuang Liu, Xin Wang, Fisher Yu, Jiashi Feng, and Trevor Darrell.
\newblock Few-shot object detection via feature reweighting.
\newblock In {\em IEEE Intl. Conf. on Computer Vision (ICCV)}, 2019.

\bibitem{repmet}
Leonid Karlinsky, Joseph Shtok, Sivan Harary, Eli Schwartz, Amit Aides, Rogerio
  Feris, Raja Giryes, and Alex~M. Bronstein.
\newblock {RepMet}: Representative-based metric learning for classification and
  few-shot object detection.
\newblock In {\em IEEE Conf. on Computer Vision and Pattern Recognition
  (CVPR)}, 2019.

\bibitem{adam}
Diederik~P. Kingma and Jimmy Ba.
\newblock Adam: {A} method for stochastic optimization.
\newblock In {\em 3rd Intl. Conf. on Learning Representations, {ICLR}}, 2015.

\bibitem{cme}
Bohao Li, Boyu Yang, Chang Liu, Feng Liu, Rongrong Ji, and Qixiang Ye.
\newblock Beyond max-margin: Class margin equilibrium for few-shot object
  detection.
\newblock In {\em {CVPR}}, June 2021.

\bibitem{local_desc}
Wenbin Li, Lei Wang, Jinglin Xu, Jing Huo, Yang Gao, and Jiebo Luo.
\newblock Revisiting local descriptor based image-to-class measure for few-shot
  learning.
\newblock In {\em IEEE Conf. on Computer Vision and Pattern Recognition
  (CVPR)}, June 2019.

\bibitem{mmfsod}
Yuewen Li, Wenquan Feng, Shuchang Lyu, Qi~Zhao, and Xuliang Li.
\newblock Mm-fsod: Meta and metric integrated few-shot object detection.
\newblock 12 2020.

\bibitem{coco}
Tsung-Yi Lin, Michael Maire, Serge Belongie, James Hays, Pietro Perona, Deva
  Ramanan, Piotr Doll{\'a}r, and C.~Lawrence Zitnick.
\newblock Microsoft {COCO:} {C}ommon objects in context.
\newblock In {\em Computer Vision -- ECCV 2014}, pages 740--755, 2014.

\bibitem{majee2021fewshot}
Anay Majee, Kshitij Agrawal, and Anbumani Subramanian.
\newblock Few-shot learning for road object detection.
\newblock In {\em AAAI Workshop on Meta-Learning and MetaDL Challenge}, volume
  140 of {\em Proceedings of Machine Learning Research}, pages 115--126, 2021.

\bibitem{fomaml}
Alex Nichol, Joshua Achiam, and John Schulman.
\newblock On first-order meta-learning algorithms.
\newblock {\em ArXiv}, abs/1803.02999, 2018.

\bibitem{opl}
Kanchana Ranasinghe, Muzammal Naseer, Munawar Hayat, Salman~H. Khan, and
  Fahad~Shahbaz Khan.
\newblock Orthogonal projection loss.
\newblock 2021.

\bibitem{metalstm}
Sachin Ravi and Hugo Larochelle.
\newblock Optimization as a model for few-shot learning.
\newblock In {\em 5th Intl. Conf. on Learning Representations, {ICLR} 2017,
  Toulon, France, April 24-26, 2017, Conference Track Proceedings}, 2017.

\bibitem{faster-rcnn}
Shaoqing Ren, Kaiming He, Ross~B. Girshick, and J.~Sun.
\newblock Faster r-cnn: Towards real-time object detection with region proposal
  networks.
\newblock {\em IEEE Transactions on Pattern Analysis and Machine Intelligence},
  2015.

\bibitem{mann}
Adam Santoro, Sergey Bartunov, Matthew Botvinick, Daan Wierstra, and Timothy
  Lillicrap.
\newblock Meta-learning with memory-augmented neural networks.
\newblock In {\em Proc. of The 33rd Intl. Conf. on Machine Learning},
  volume~48, pages 1842--1850, 2016.

\bibitem{fsgraph}
Victor~Garcia Satorras and Joan~Bruna Estrach.
\newblock Few-shot learning with graph neural networks.
\newblock In {\em Intl. Conf. on Learning Representations}, 2018.

\bibitem{protonet}
Jake Snell, Kevin Swersky, and Richard Zemel.
\newblock Prototypical networks for few-shot learning.
\newblock In {\em Advances in Neural Information Processing Systems}, pages
  4077--4087, 2017.

\bibitem{fscontrastive}
Bo~Sun, Banghuai Li, Shengcai Cai, Ye~Yuan, and Chi Zhang.
\newblock Fsce: Few-shot object detection via contrastive proposal encoding.
\newblock In {\em Proc. of the IEEE conf. on computer vision and pattern
  recognition (CVPR)}, June 2021.

\bibitem{relation-net}
Flood Sung, Yongxin Yang, Li~Zhang, Tao Xiang, Philip~H.S. Torr, and Timothy~M.
  Hospedales.
\newblock Learning to {Compare:} {R}elation network for few-shot learning.
\newblock In {\em IEEE Conf. on Computer Vision and Pattern Recognition
  (CVPR)}, June 2018.

\bibitem{idd}
G.~{Varma}, A.~{Subramanian}, A.~{Namboodiri}, M.~{Chandraker}, and C.~V.
  {Jawahar}.
\newblock {IDD:} {A} dataset for exploring problems of autonomous navigation in
  unconstrained environments.
\newblock In {\em IEEE Winter Conf. on Applications of Computer Vision (WACV)},
  pages 1743--1751, 2019.

\bibitem{matching-net}
Oriol Vinyals, Charles Blundell, Timothy Lillicrap, koray kavukcuoglu, and Daan
  Wierstra.
\newblock Matching networks for one shot learning.
\newblock In {\em Advances in Neural Information Processing Systems}, 2016.

\bibitem{mm_optim}
Duo Wang, Yu~Cheng, Mo~Yu, Xiaoxiao Guo, and Tao Zhang.
\newblock A hybrid approach with optimization-based and metric-based
  meta-learner for few-shot learning.
\newblock {\em Neurocomputing}, 349:202--211, 2019.

\bibitem{fsdet}
Xin Wang, Thomas~E. Huang, Trevor Darrell, Joseph~E Gonzalez, and Fisher Yu.
\newblock Frustratingly simple few-shot object detection.
\newblock In {\em Intl. Conf. on Machine Learning (ICML)}, 2020.

\bibitem{metadet}
Yu-Xiong Wang, Deva Ramanan, and Martial Hebert.
\newblock Meta-learning to detect rare objects.
\newblock In {\em IEEE Intl. Conf. on Computer Vision (ICCV)}, 2019.

\bibitem{addfeat}
Yang Xiao and Renaud Marlet.
\newblock Few-{S}hot object detection and viewpoint estimation for objects in
  the wild.
\newblock In {\em European Conf. on Computer Vision (ECCV)}, 2020.

\bibitem{metarcnn}
Xiaopeng Yan, Ziliang Chen, Anni Xu, Xiaoxi Wang, Xiaodan Liang, and Liang Lin.
\newblock {Meta R-CNN:} {T}owards general solver for instance-level low-shot
  learning.
\newblock In {\em IEEE Conf. on Computer Vision and Pattern Recognition
  (CVPR)}, pages 9577--9586, 2019.

\bibitem{pnpdet}
Gongjie Zhang, Kaiwen Cui, Rongliang Wu, Shijian Lu, and Yonghong Tian.
\newblock {PNPDet:} {E}fficient {Few-Shot} detection without forgetting via
  {Plug-and-Play} sub-networks.
\newblock In {\em Proc. of the IEEE/CVF Winter Conf. on Appl. of Computer
  Vision (WACV)}, pages 3823--3832, 2021.

\bibitem{metadetr}
Gongjie Zhang, Zhipeng Luo, Kaiwen Cui, and Shijian Lu.
\newblock Meta-detr: Few-shot object detection via unified image-level
  meta-learning, 2021.

\end{thebibliography}
\bibliographystyle{plain}

\newpage
\appendix

\section{Appendix}

\subsection{Architecture of The Split and Excite Module}
\begin{figure}[h] 
    \centering
    \includegraphics[width=0.7\textwidth]{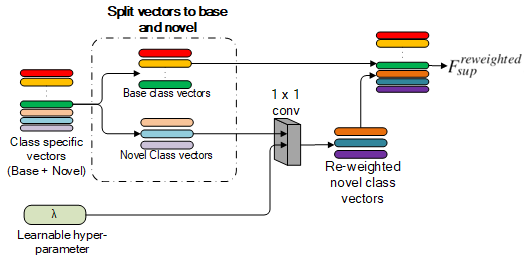}
    \caption{Architecture of the proposed Split and Excite module.}
     \label{fig:split_excite}
\end{figure}
Following the definition of the Split and Excite (SE) module in section \ref{meth:split_excite} the sub-network can be decomposed into three distinct parts. Figure \ref{fig:split_excite} demonstrates the detailed architecture of the Split and Excite module.
The output of the meta branch is a set of class specific attention vectors $F_{sup}$. 
At first, we segregate the class attention vectors into base and novel classes. 
Secondly, we pass the attention vectors belonging to the novel classes and the learnable hyper-parameter $\lambda$ through a $1 \times 1$ convolution layer which re-weights the novel class vectors higher than the base classes, $\lambda > 1$. Finally, we concatenate the re-weighted feature vectors of the novel classes with those of the base classes to produce $F_{sup}^{reweighted}$.

\subsection{Additional Qualitative Results on IDD}
\newcommand{\centered}[1]{\begin{tabular}{l} #1 \end{tabular}}

\begin{figure*}[tp]
        \centering
        \begin{tabular}{ccc}
                \multicolumn{3}{c}{Positive results from novel classes in IDD-OS for 10-shot setting.} \\
                \includegraphics[width=0.33\textwidth]{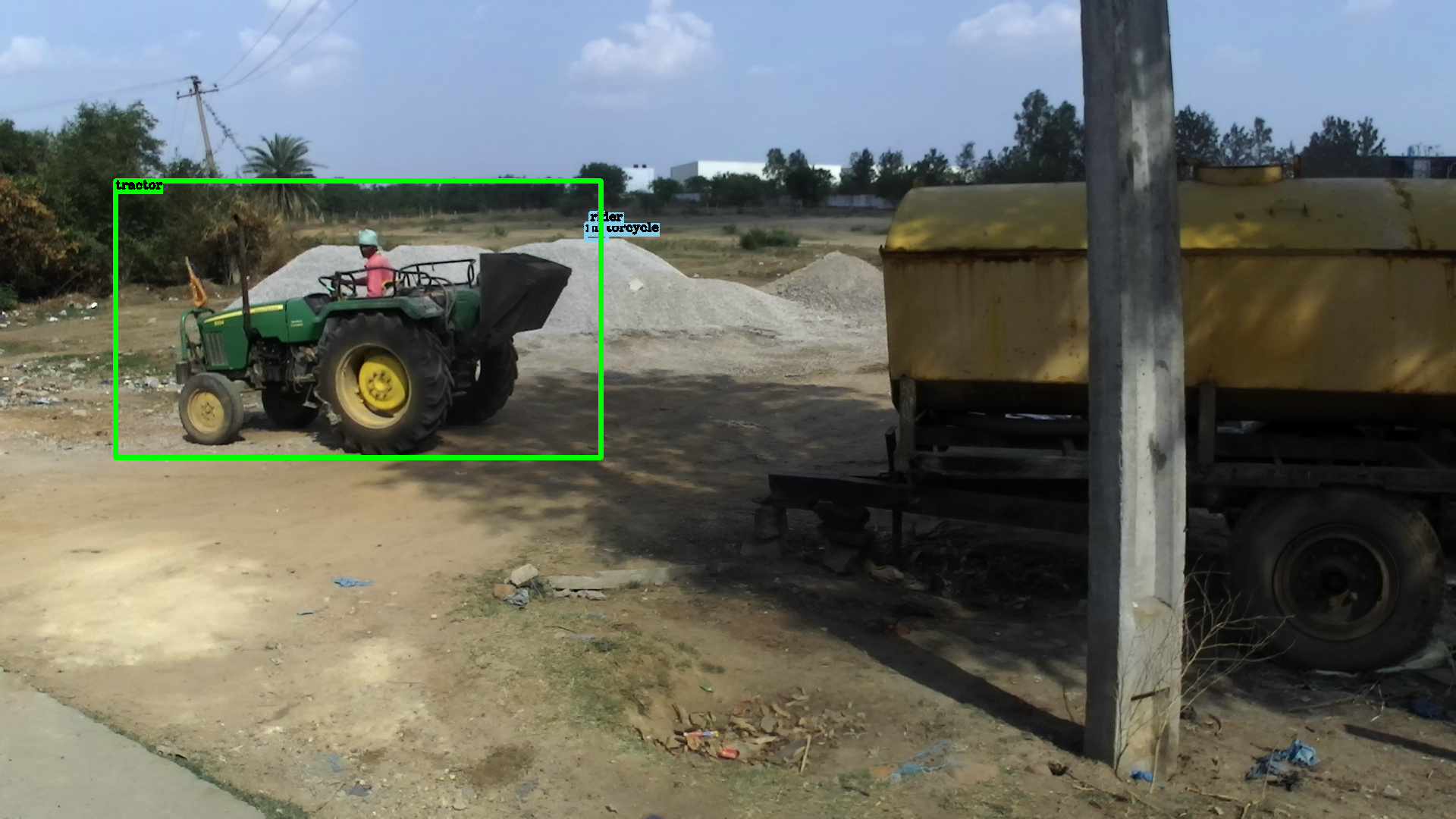} & \includegraphics[width=0.33\textwidth]{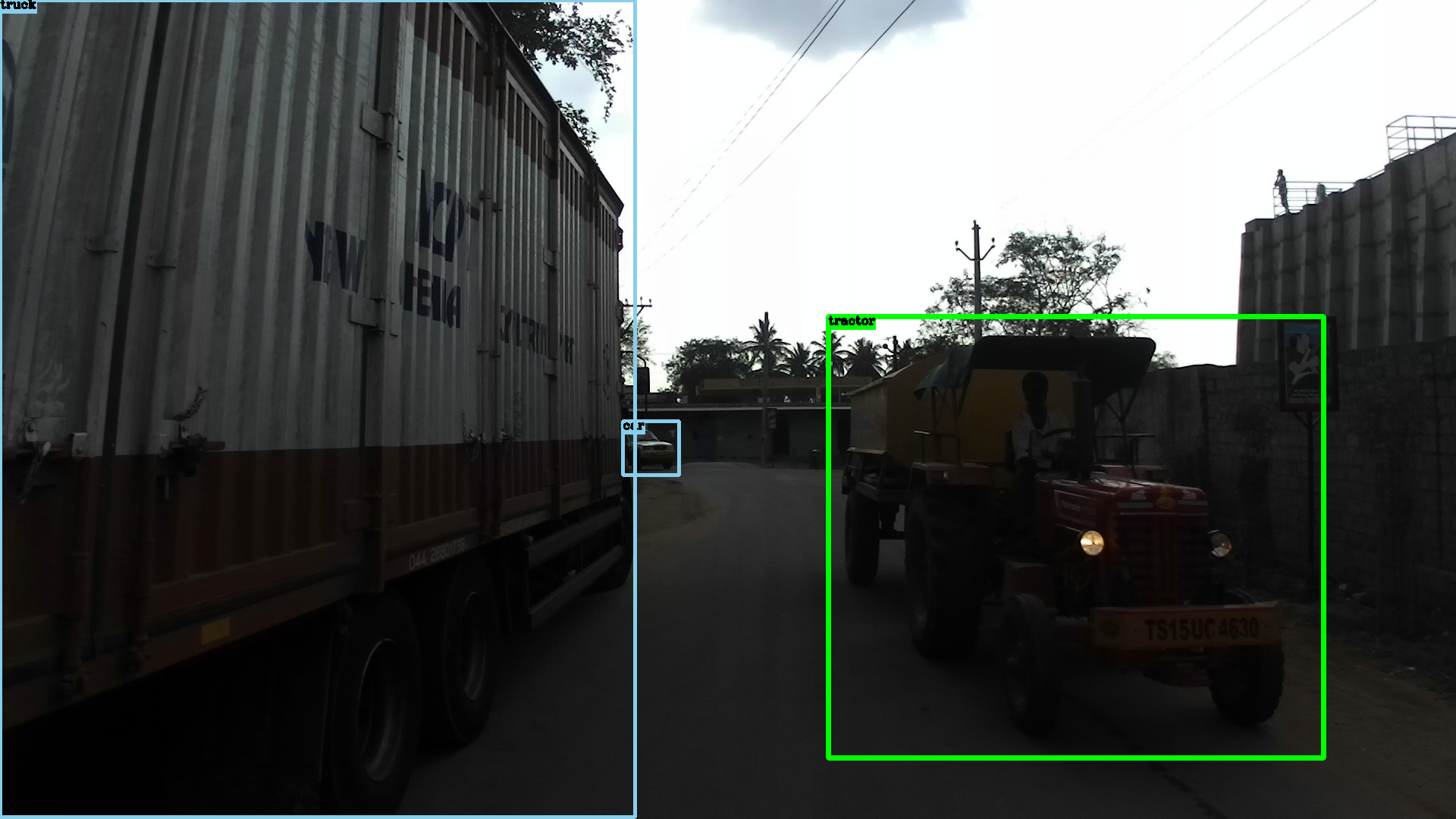} & \includegraphics[width=0.33\textwidth]{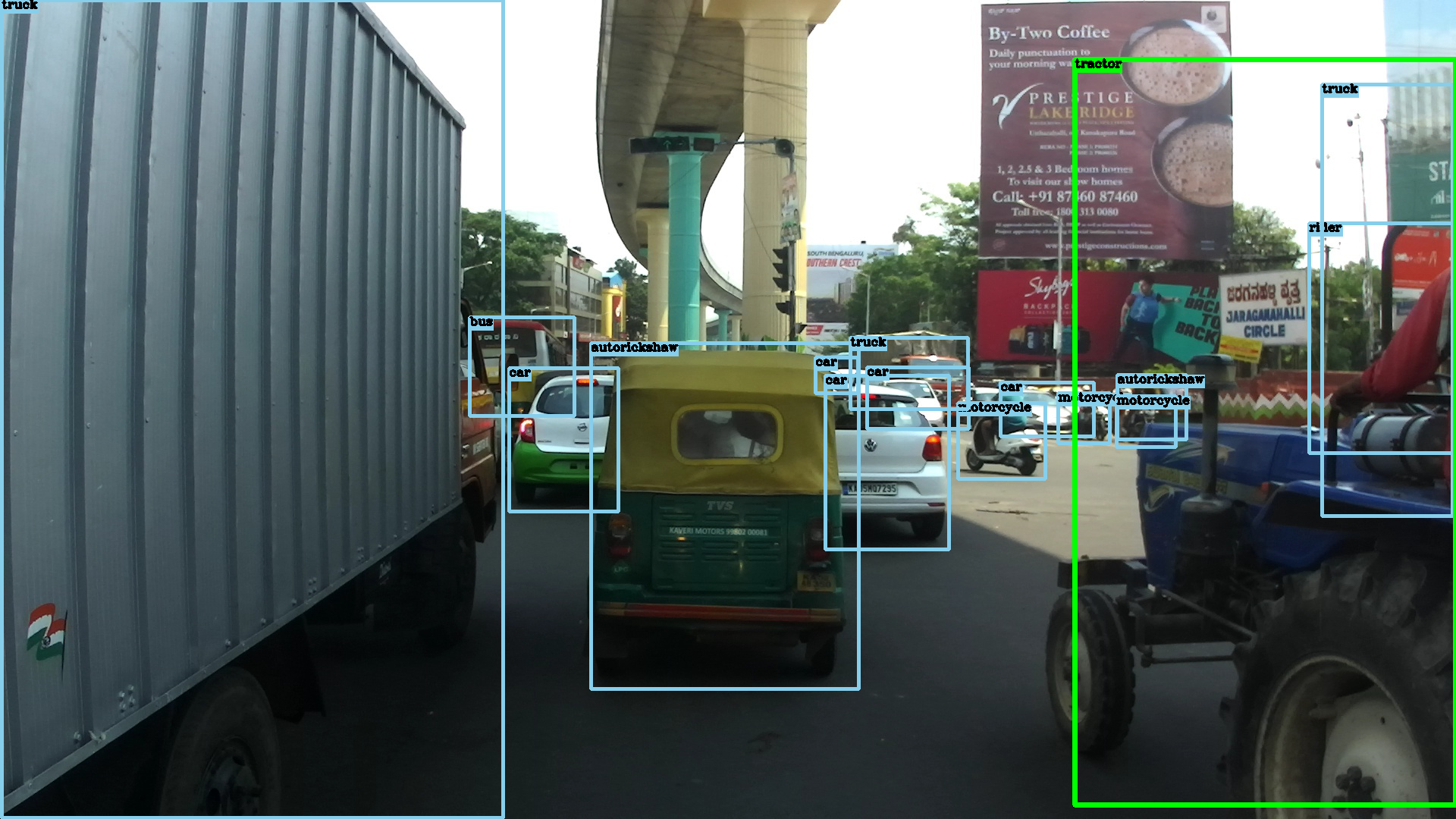} \\ \includegraphics[width=0.33\textwidth]{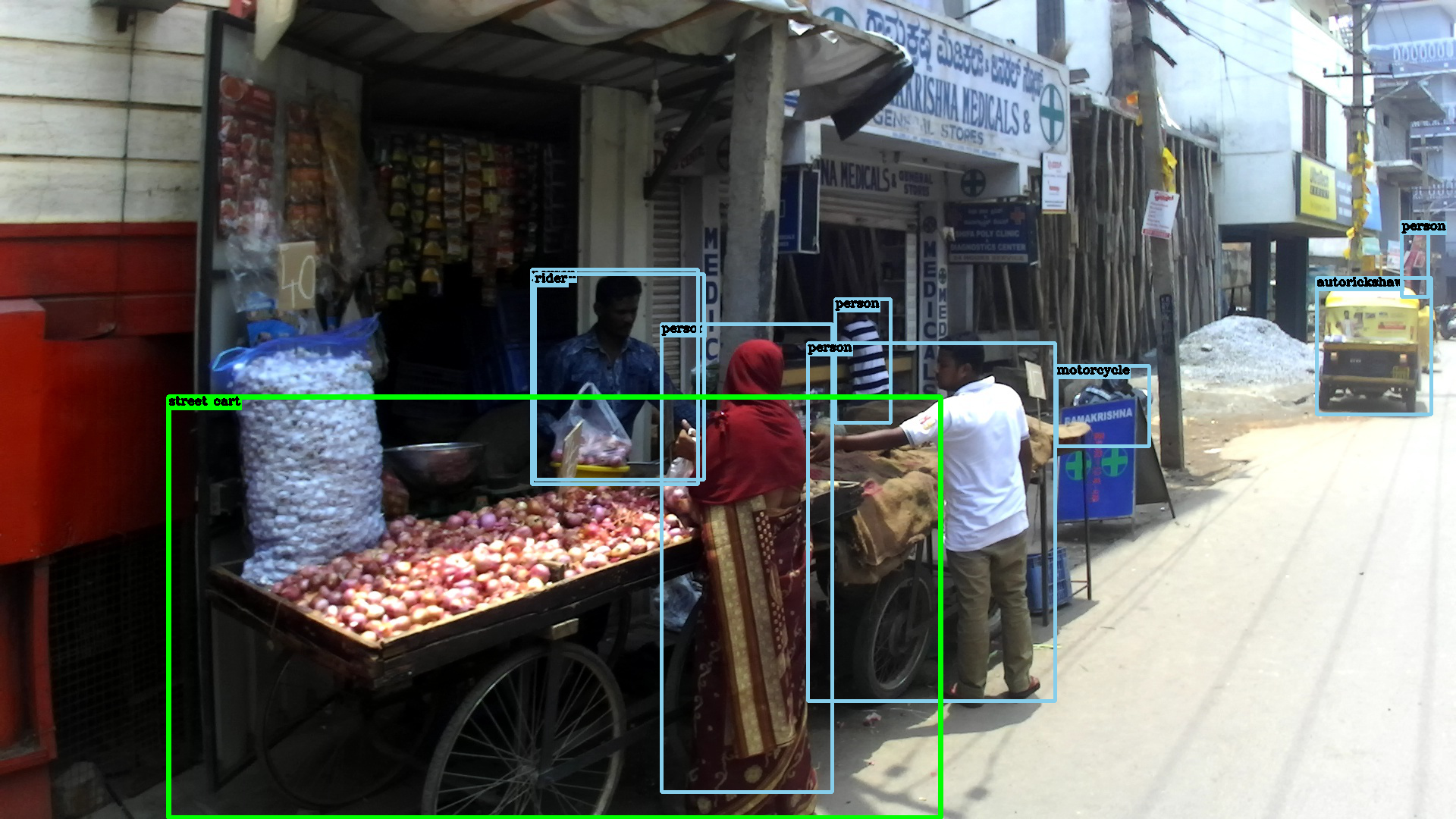}  &
                \includegraphics[width=0.33\textwidth]{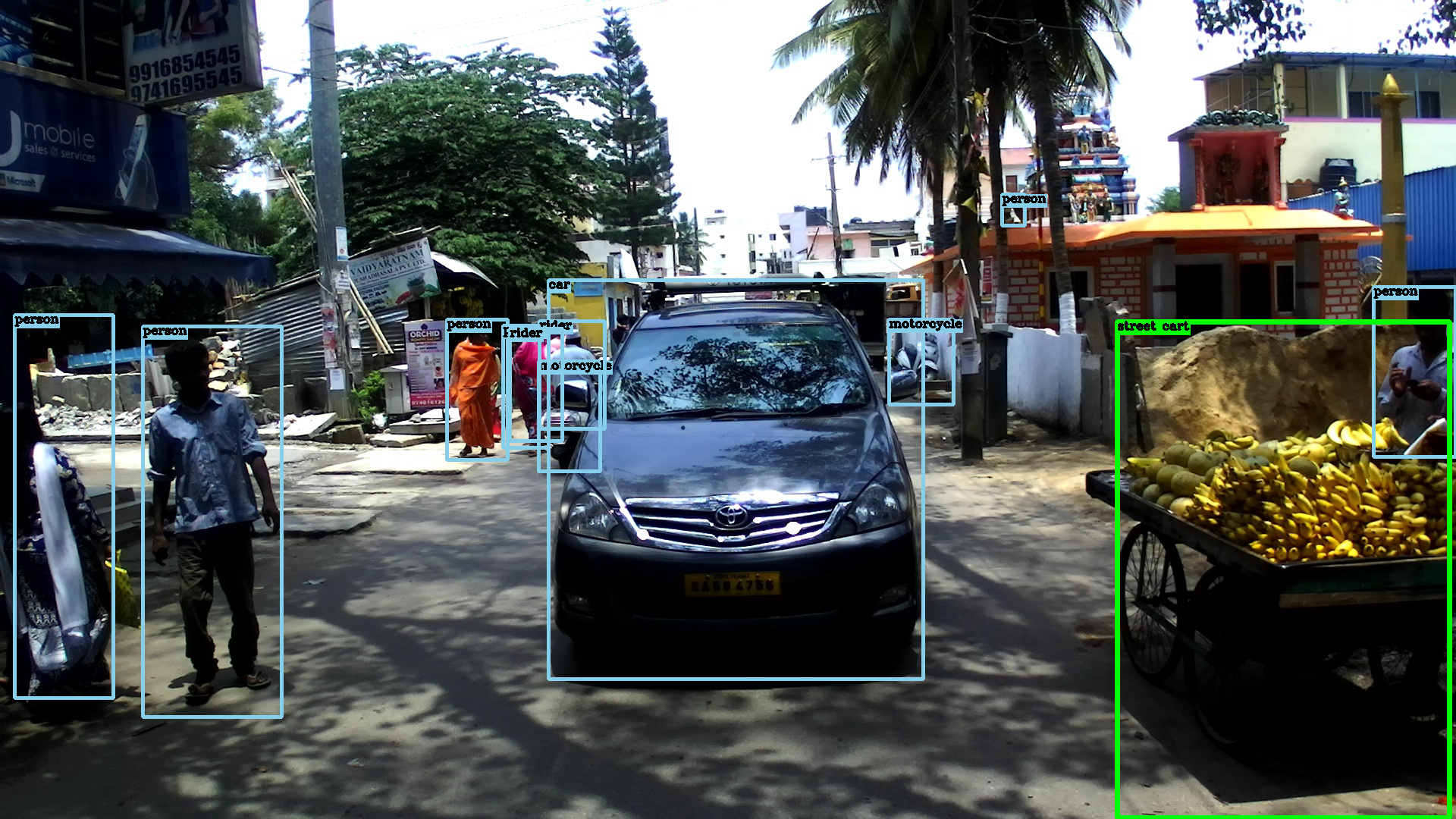} & \includegraphics[width=0.33\textwidth]{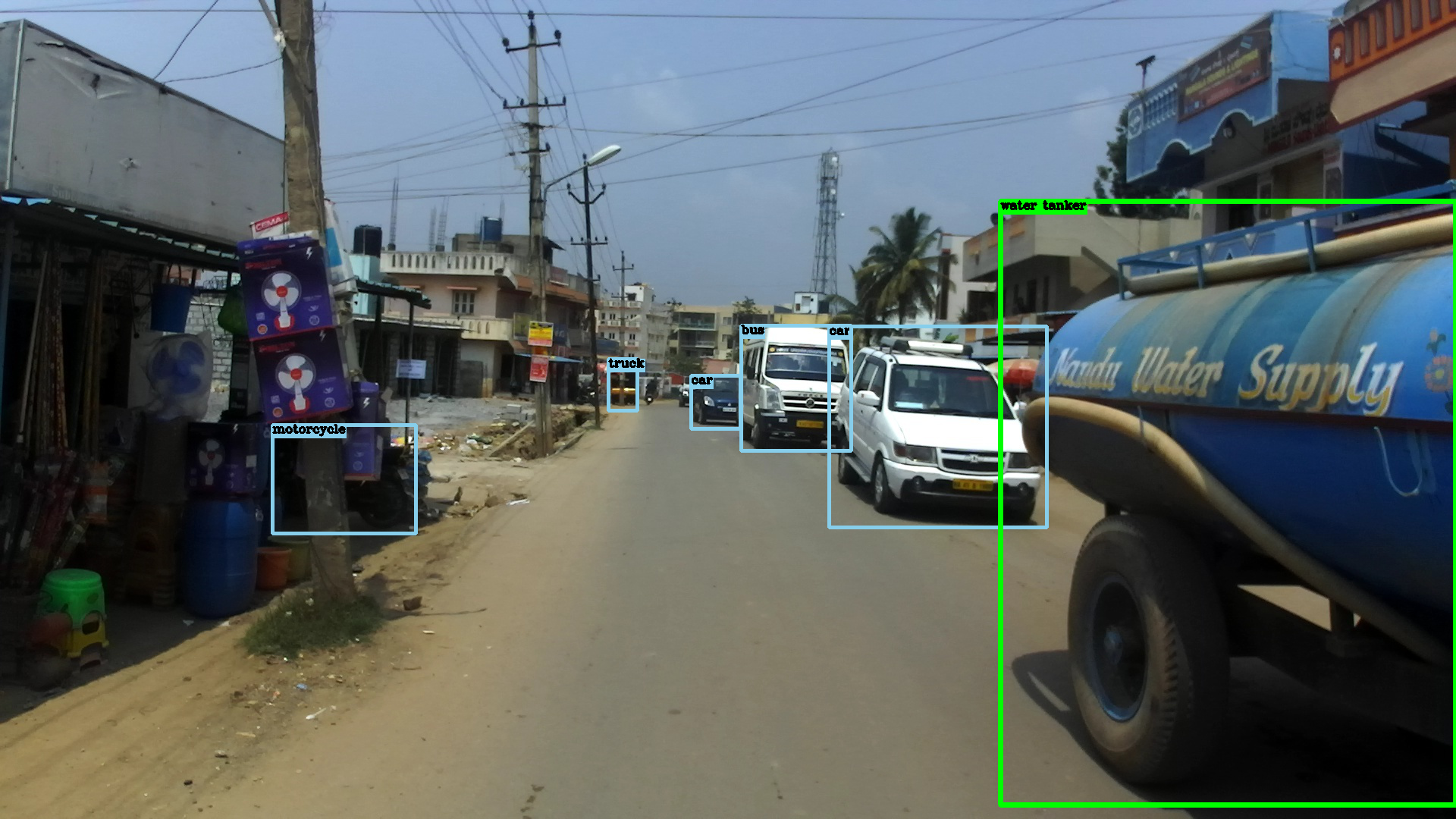} \\ \includegraphics[width=0.33\textwidth]{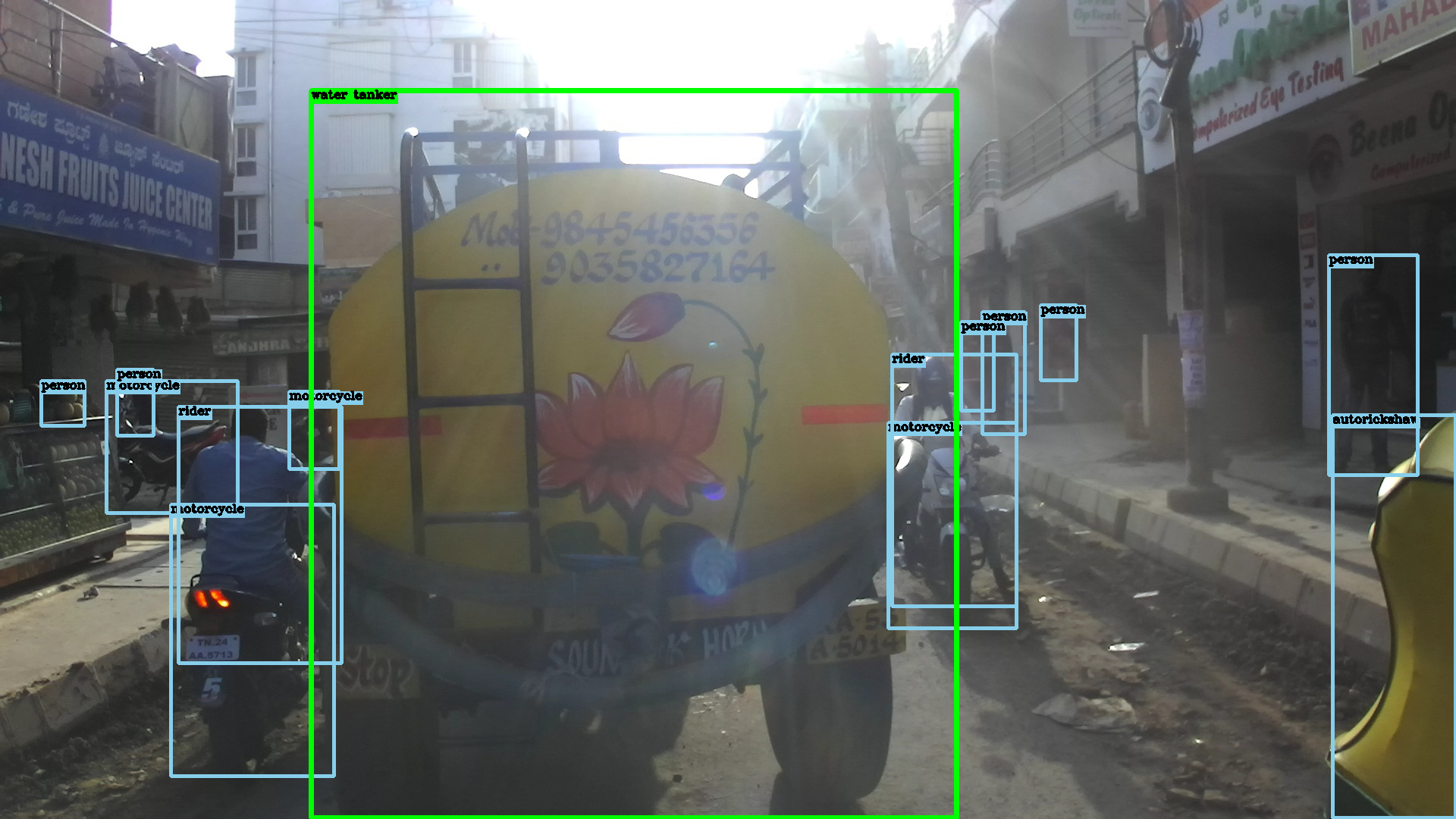} & \includegraphics[width=0.33\textwidth]{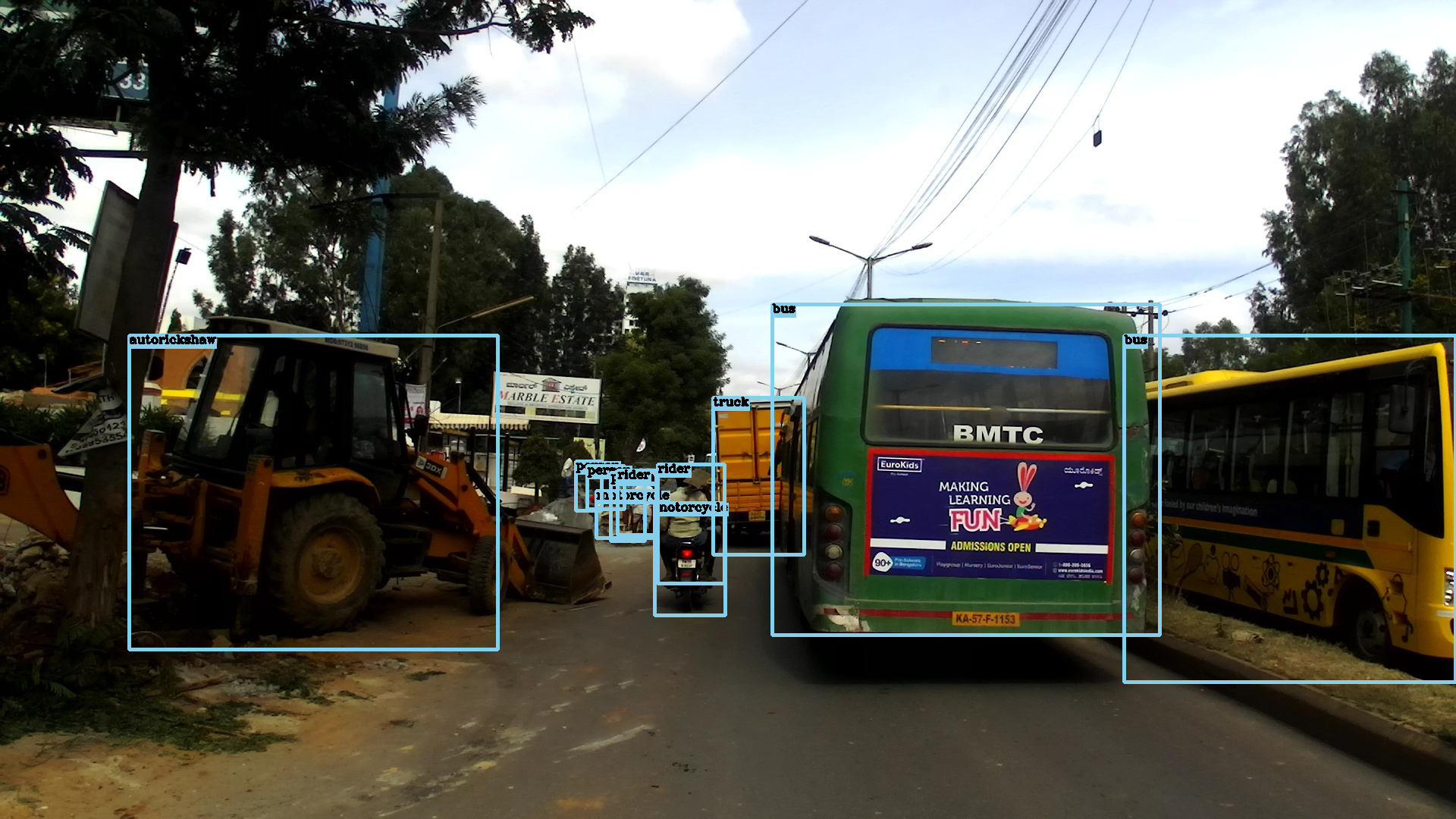} &
                \includegraphics[width=0.33\textwidth]{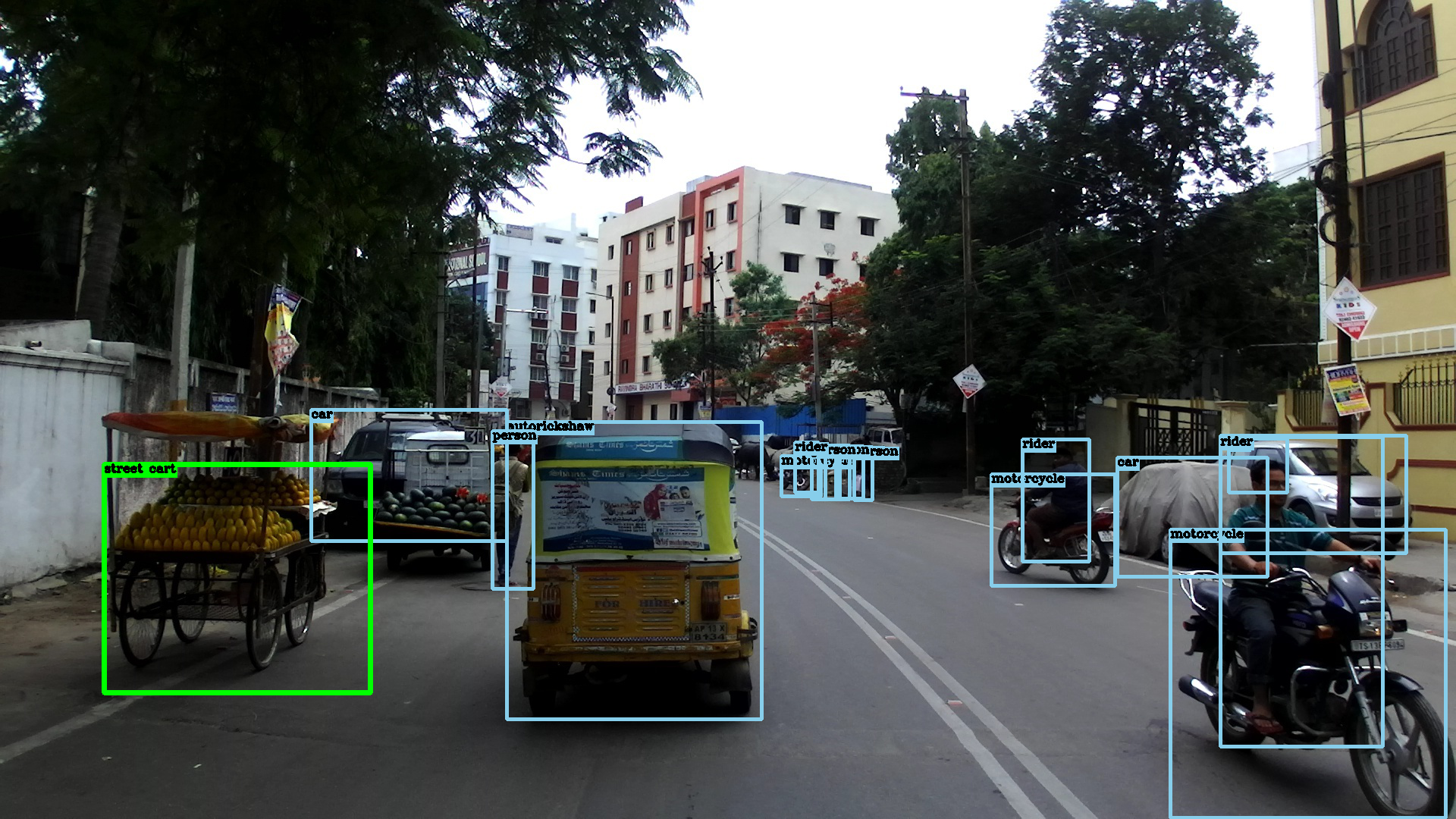} \\
                \multicolumn{3}{c}{Negative predictions on IDD-OS split in 10-shot setting.} \\
                \includegraphics[width=0.33\textwidth]{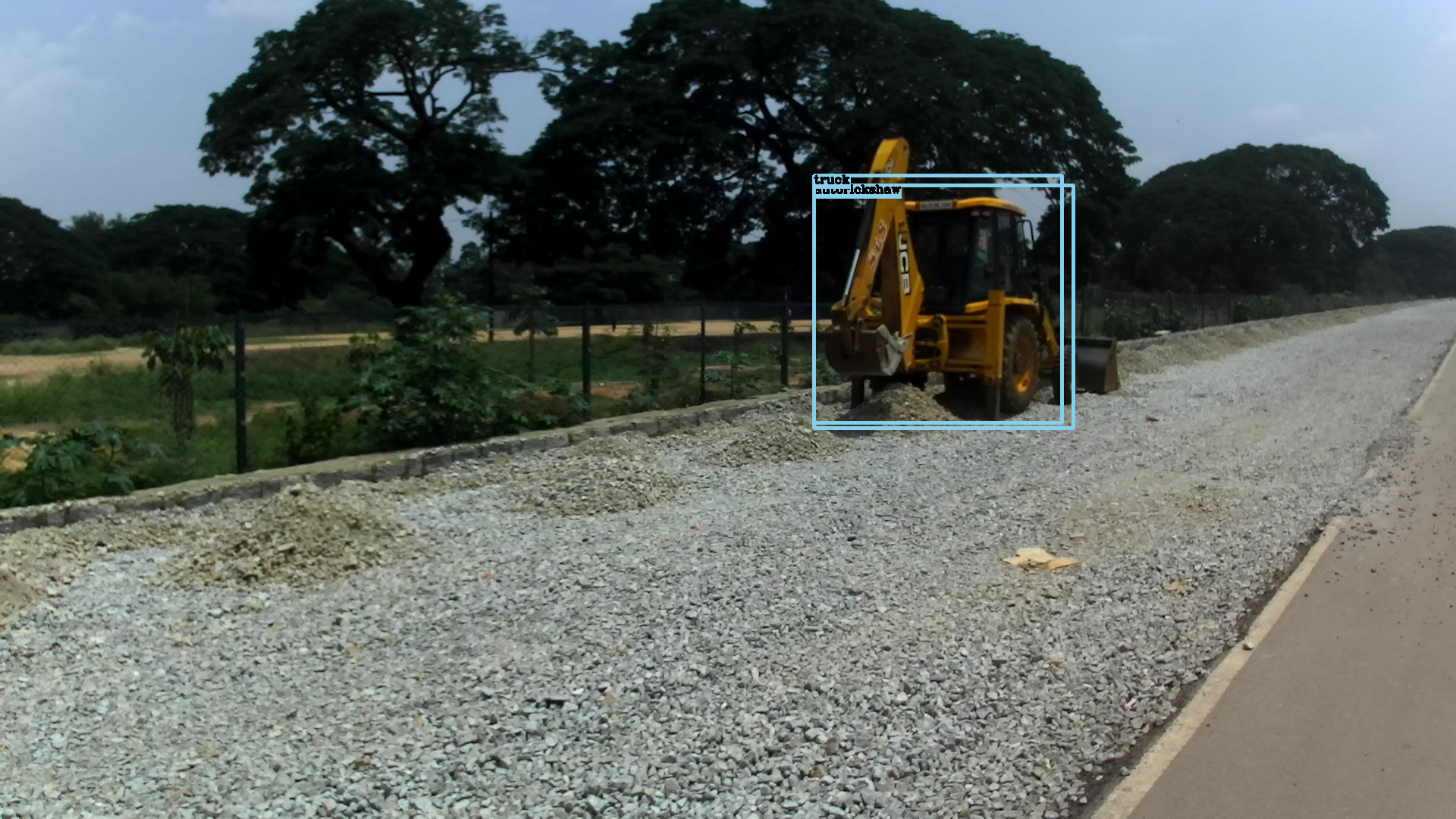} & \includegraphics[width=0.33\textwidth]{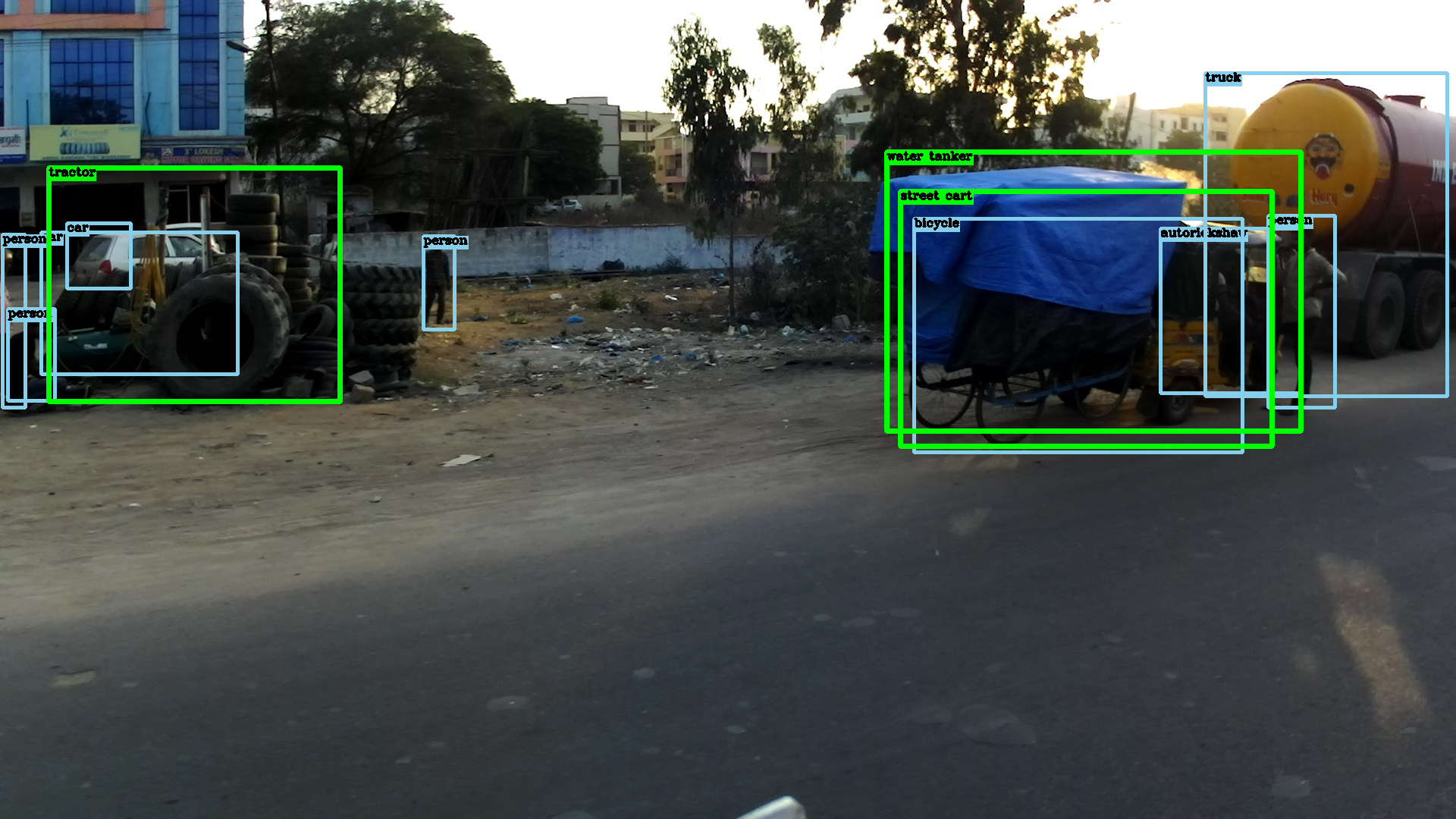} & \includegraphics[width=0.33\textwidth]{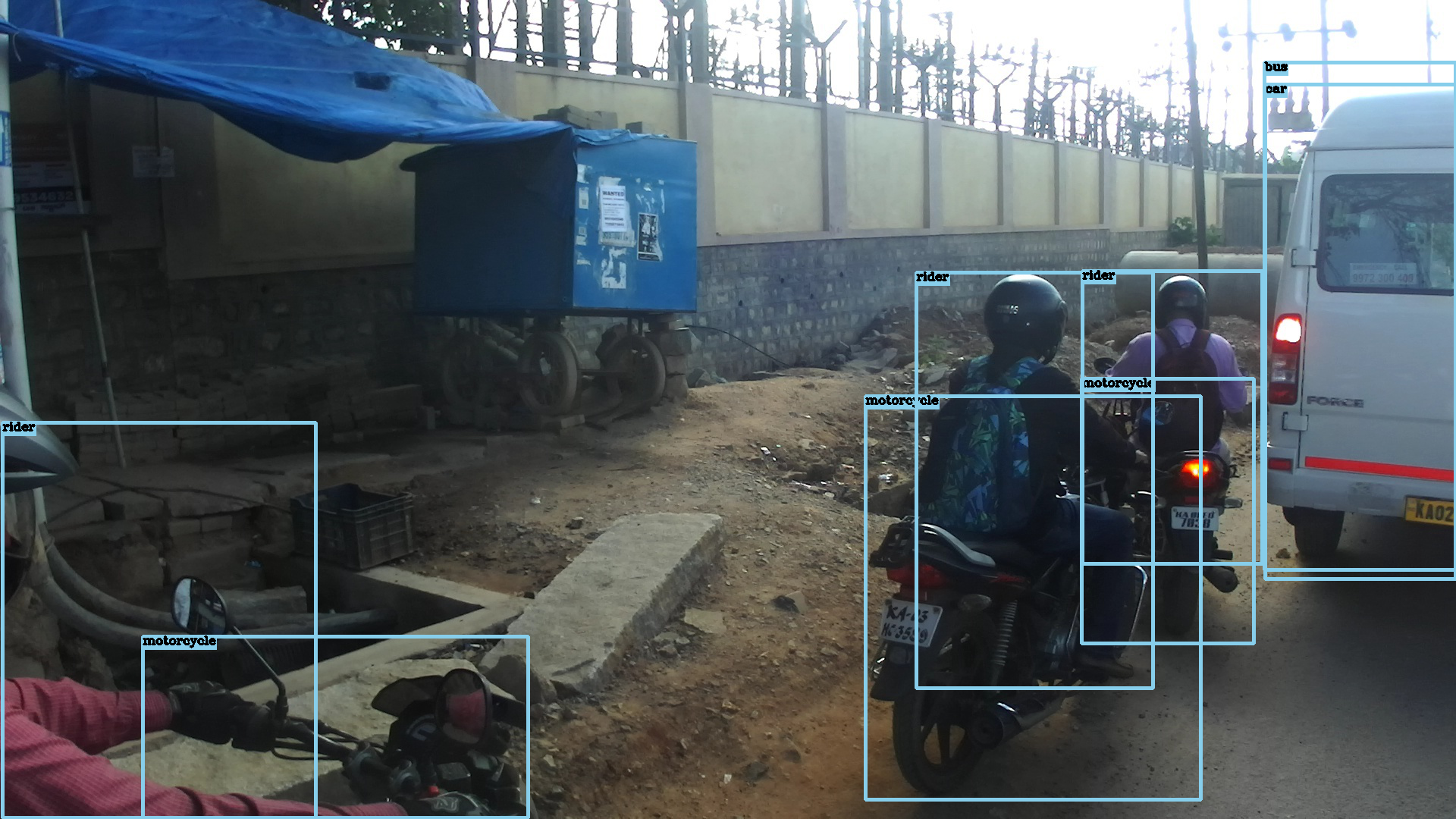}
        \end{tabular}
        \caption{Qualitative results from the MGML approach on the few-shot India Driving Dataset (IDD-OS split) for the 10-shot setting. The first three rows in the figure show positive predictions while the final row shows failure cases where we continue to observe class confusion and catastrophic forgetting.}
        \label{fig:add_qual}
\end{figure*}
Figure \ref{fig:add_qual} demonstrates addiitonal qualitative results from the MGML approach on the novel classes in IDD-OS split of the India Driving Dataset (IDD) in the 10-shot setting. The positive results indicates that our method is resistant to the major pitfalls in standard object detection algorithms - occlusions, variational lighting etc. It also points out the reduction in class confusion between co-occuring and visually similar classes such as \emph{street cart} and \emph{bicycle} (visually similar), or \emph{motorcycle} and \emph{rider} (co-occuring). The negative results indicate the existence of class confusion in object classes which have large visual similarity with existing classes such as \emph{Excavator (JCB)}. Such issues will be addressed in future research. 

\end{document}